%% file: ms.tex
\documentclass[10pt,journal,compsoc]{IEEEtran}

\usepackage{adjustbox}
\usepackage{booktabs}
\ifCLASSOPTIONcompsoc
  \usepackage[nocompress]{cite}
\else
  \usepackage{cite}
\fi
\ifCLASSINFOpdf
\else
\fi
\usepackage{color}

\newcommand{\comment}[1]{}

\newcommand{\minunder}{\mathop{\mathrm{min}}\limits}
\newcommand{\maxunder}{\mathop{\mathrm{max}}\limits}

\newcommand{\argmin}{\mathop{\mathrm{arg min}}\limits}

\usepackage{amssymb}
\usepackage{multirow}
\usepackage{booktabs}
\usepackage{mathtools}
\usepackage{subcaption}
\usepackage{dblfloatfix}
\usepackage[table,x11names]{xcolor}
\usepackage{comment}

\newcommand{\mathbbm}[1]{\text{\usefont{U}{bbm}{m}{n}#1}}

\begin{document}
\title{DPODv2: Dense Correspondence-Based 6 DoF Pose Estimation}

\author{Ivan~Shugurov,
        Sergey~Zakharov,
        Slobodan~Ilic% <-this % stops a space

\IEEEcompsocitemizethanks{\IEEEcompsocthanksitem I. Shugurov, and S. Ilic, are with the Department of Informatics, Technical University of Munich, Germany and with Siemens AG, Munich, Germany.\protect\\
E-mail: see http://campar.in.tum.de
\IEEEcompsocthanksitem S. Zakharov was affiliated with the Department of Informatics, Technical University of Munich and Siemens AG at the time of the trial and is currently affiliated with Toyota Research Institute, Los Altos, USA\protect\\
E-mail: sergey.zakharov@tri.global
}% <-this % stops an unwanted space

\thanks{Manuscript received May 23, 2021.}}


\markboth{IEEE TRANSACTIONS ON PATTERN ANALYSIS AND MACHINE INTELLIGENCEE, ARXIV PREPRINT}%
{Shell \MakeLowercase{\textit{et al.}}: Bare Demo of IEEEtran.cls for Computer Society Journals}
\IEEEtitleabstractindextext{%
\begin{abstract}
We propose a three-stage 6 DoF object detection method called DPODv2 (Dense Pose Object Detector) that relies on dense correspondences. % pipeline based on the recent advances in the field. 
We combine a 2D object detector with a dense correspondence estimation network and a multi-view pose refinement method to estimate a full 6 DoF pose. Unlike other deep learning methods that are typically restricted to monocular RGB images, we propose a unified deep learning network allowing different imaging modalities to be used (RGB or Depth). Moreover, we propose a novel pose refinement method, that is based on differentiable rendering. The main concept is to compare predicted and rendered correspondences in multiple views to obtain a pose which is consistent with predicted correspondences in all views. Our proposed method is evaluated rigorously on different data modalities and types of training data in a controlled setup. The main conclusions is that RGB excels in correspondence estimation, while depth contributes to the pose accuracy if good 3D-3D correspondences are available. Naturally, their combination achieves the overall best performance. 
We perform an extensive evaluation and an ablation study to analyze and validate the results on several challenging datasets. DPODv2 achieves excellent results on all of them while still remaining fast and scalable independent of the used data modality and the type of training data.
\end{abstract}

\begin{IEEEkeywords}
6 DoF pose estimation, dense correspondences, synthetic data.
\end{IEEEkeywords}}

\maketitle

\IEEEdisplaynontitleabstractindextext

% \fi
\IEEEpeerreviewmaketitle

\input{introduction}

\input{related}

\input{method}
\input{data_preparation}

\input{experiments}

\input{conclusion}
\input{supp_arxiv}

% or


%\appendices

%\section{}

%\ifCLASSOPTIONcompsoc
%\else
%\fi


\ifCLASSOPTIONcaptionsoff
  \newpage
\fi



%\bibliographystyle{IEEEtran}
\bibliographystyle{IEEEtran}
\bibliography{egbib}


\begin{IEEEbiography}[{\includegraphics[width=1in,height=1.25in,clip,keepaspectratio]{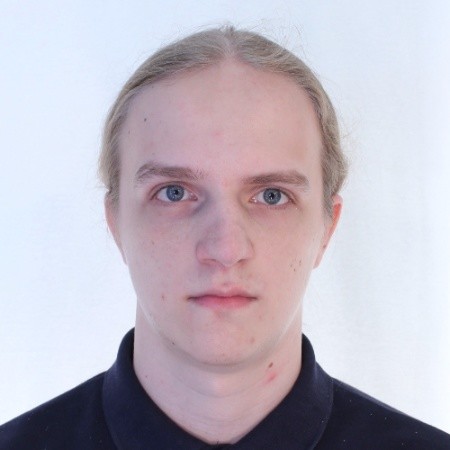}}]{Ivan Shugurov}
Is a PhD student at the Chair for Computer Aided Medical Procedures / TU Munich and at Siemens Corporate Technology under supervision of PD. Dr. Slobodan Ilic. He completed his Bachelor's degree in Software Engineering at Higher School of Economics. Subsequently, he studied Informatics at TU Munich. His research focuses on multimodal 2D/3D object detection and pose estimation.
\end{IEEEbiography}

\begin{IEEEbiography}[{\includegraphics[width=1in,height=1.25in,clip,keepaspectratio]{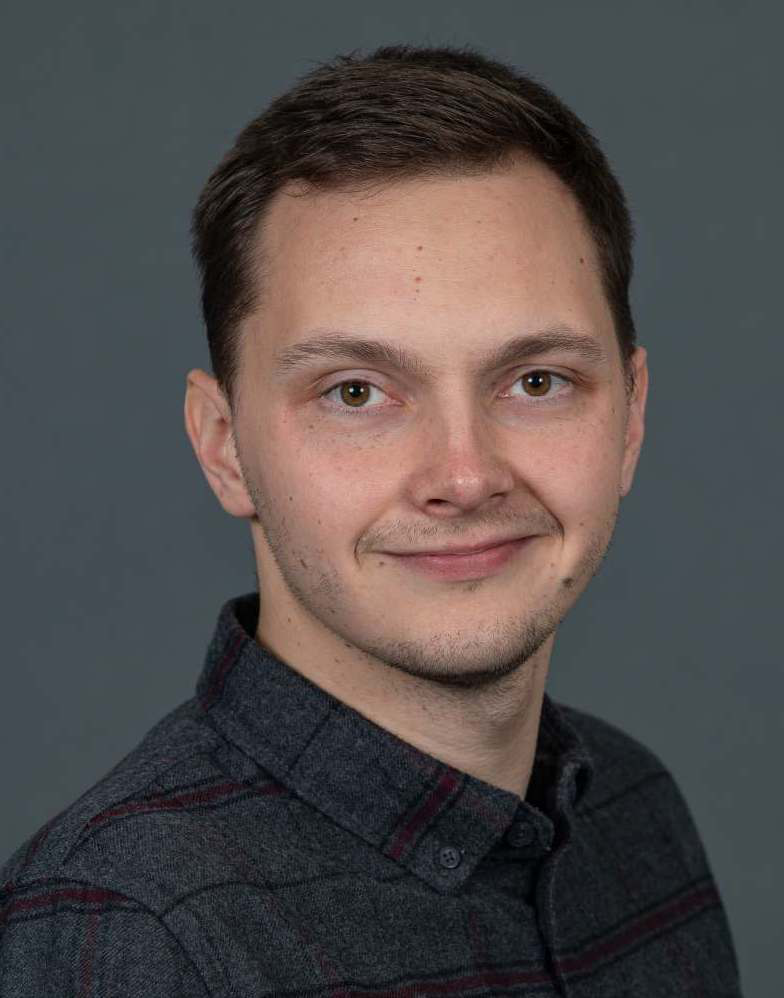}}]{Sergey~Zakharov}
is a research scientist at Toyota Research Institute (TRI) in Los Altos, CA. He obtained his MSc degree in Computational Science and Engineering and a MSc degree in Biomedical Computing from the Technical University of Munich (TUM) in 2015 and 2016 respectively. Subsequently, he completed his PhD degree at the Chair for Computer Aided Medical Procedures, TUM and Siemens Corporate Technology. His research focuses on scene understanding and covers such topics as scalable 3D object detection, neural implicit representations, differentiable rendering, and domain adaptation. He owns a number of industry-relevant solutions published and presented at first-tier computer vision and robotics conferences including CVPR, ICCV, ICRA, and IROS.
\end{IEEEbiography}
%\newpage

\begin{IEEEbiography}[{\includegraphics[width=1in,height=1.25in,clip,keepaspectratio]{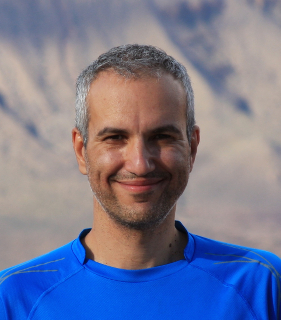}}]{Slobodan Ilic}
is currently senior key expert research scientist at Siemens Corporate Technology in Munich, Perlach. He is also a visiting researcher and lecturer at Computer Science Department of TUM and closely works with the CAMP Chair. From 2009 until end of 2013 he was leading the Computer Vision Group of CAMP at TUM, and before that he was a senior researcher at Deutsche Telekom Laboratories in Berlin. In 2005 he obtained his PhD at EPFL in Switzerland under supervision of Pascal Fua. His research interests include: 3D reconstruction, deformable surface modelling and tracking, real-time object detection and tracking, human pose estimation and semantic segmentation.
\end{IEEEbiography}


\vfill


\end{document}

%% file: introduction.tex
\IEEEraisesectionheading{\section{Introduction}\label{sec:introduction}}

Object detection and 6 DoF pose estimation are not new fields in computer vision. In fact, they are among the major driving forces, being crucial for various application fields, such as augmented reality, robotics and autonomous driving. Therefore, there is a vast number of methods trying to tackle this problem. 

In the pre-deep learning era, pose estimation was typically performed either completely using depth or using a combination of depth and RGB data. However, recent trends in 6 DoF pose estimation move towards disregarding depth data as a modality and using only RGB data. The reasons for that are manifold, but the most important are the omnipresent availability of RGB cameras in the modern devices and the success of deep learning methods. The progress in RGB deep learning allows the researchers to choose from a vast number of ready to use network architectures and focus their attention on extending them to predict the pose. The state of the art RGB solutions are solely based on convolutional neural networks (CNNs), demonstrating impressive results that could barely be imaginable a couple of years ago. Even though recent RGB methods, such as CosyPose~\cite{labbe2020cosypose} perform better than depth-based point pair feature approaches~\cite{drost2010model}, they still suffer from perspective ambiguities and appearance changes. This can be remedied with the
additional use of depth data or an extra step with multi-view refinement.

Inspired by the works of Taylor et al.~\cite{taylor2012vitruvian} and Gueler et al.~\cite{gulerDenseposeDenseHuman2018}, Brachmann et al~\cite{brachmann2014learning} and Jafari et al.~\cite{jafari2018ipose}, we developed and compared several variations of deep dense correspondence-based 6 DoF object detectors using either RGB or depth as input. We ran a thorough analysis and ablation studies to evaluate the strengths and weaknesses of the presented modalities and compared them to the state of the art. Moreover, each correspondence estimation networks is trained on two types of data: synthetic and real. While real labeled data are most commonly used to achieve the best possible results, the acquisition of such data is often infeasible in real applications due to the high costs and significant time efforts. Synthetic data, on the other hand, is free of these drawbacks and can be easily rendered in a variety of scenes and under an unlimited number of poses. Unfortunately, the methods trained on synthetic data are subject to the domain gap problem, due to the dissimilarity between real and synthetic images. This might causes them to perform worse than the methods trained on real data. However, as was shown in HomebrewedDB~\cite{kaskman2019homebreweddb},  networks trained only synthetic data are less prone to overfitting due to a larger data variability, which allows them to perform similarly to the methods trained on real data when train and test images do not come from the same image sequence. Therefore, detectors trained on synthetic data are desirable, because the required data is easier to obtain and because such detectors tend to generalize better, thus being of higher practical significance.

\begin{figure*}
    \centering
    \includegraphics[width=1\linewidth]{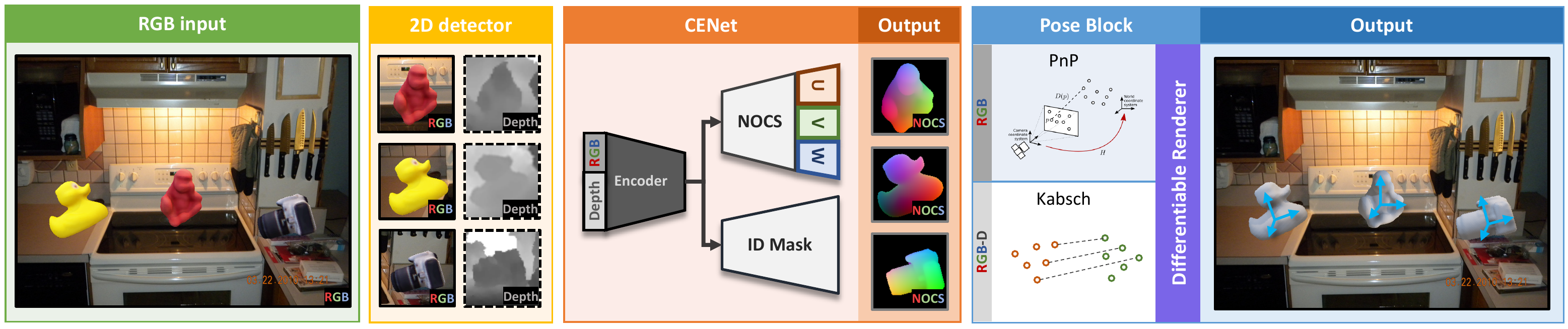}
    \caption{\textbf{Synthetic toy example illustrating three-stage correspondence-based 6 DoF pose estimation:} A full RGB image is fed to a 2D object detector for bounding box estimation. Resulting bounding boxes are then used to generate crops on the available data, which are subsequently fed into the correspondence network. If only RGB images are provided, then the pose is estimated from correspondences using 2D-3D PnP. In case registered depth data is also in place, we project the estimated correspondences into 3D and use the 3D-3D Kabsch algorithm.}
    \label{fig:pipeline}
    \vspace{-1.5em}
\end{figure*}

Even though there is a larger number of pose estimation methods for all kinds of data modalities, they are typically designed for a specific one and have completely different architectures from methods operating on other data modalities. This makes it impossible to measure which advances come from better methods and which come from a different data modality or from better training data. In this paper, we propose a unified deep neural network architecture capable of predicting dense correspondences either from depth maps or from RGB images. The pose estimation part enables direct comparison of the data modalities on various datasets. Additionally, in contrast to the other dense correspondence methods, we report per-pixel correspondence error to deepen the understanding of where pose imprecision comes from. From the results obtained we conclude that RGB images excel in object localization and correspondence estimation, while 3D-3D correspondences between the depth images and the model result in more accurate pose estimates. Naturally, their combination brings the best performance. We also introduce a correspondence-based refinement method based on differentable rendering.  Moreover, if relative camera transformations between frames are known, the predicted poses can be additionally refined by aligning per-image dense correspondences between the object correspondences rendered in the predicted pose and the correspondences predicted by the network. The proposed loss function is implemented using a differentible renderer, which allows for its minimization with the standard gradient-based methods.
    
The main contributions of this work can be summarized as follows:
\begin{itemize}
    \item A unified framework for dense correspondence estimation, whose architecture is agnostic to the input modality type available (RGB or Depth);
    \item A thorough analysis and ablation study of the presented methods and estimation of their strengths and weaknesses;
    \item An analysis of the quality of predicted correspondences;
    \item A correspondence-based 6 DoF pose refinement extensible to multiple views.
\end{itemize}

%% file: related.tex
\section{Related Work}\label{sec:related_work}

The topic of 6 DoF pose estimation has been studied extensively of times in the past. In this section, we provide a quick overview over the most prominent works. We coarsely grouped the existing 6 DoF pose estimation methods into three categories: direct pose regression, template-based and correspondence-based  methods. We additionally discuss multi-view methods for pose estimation and refinement methods.

\subsection{Direct Pose Regression Methods}

Given that neural networks are universal approximators, it feels natural to try to tackle the problem of pose estimation by direct prediction of the pose. One of the first works to ever do that was the PoseNet~\cite{kendall2015posenet}, where a CNN is utilized to regress the position and orientation of a camera given an RGB image. Subsequent works improved the localization results by combining a CNN with LSTM units~\cite{walch2017image} and studied the effect of different loss functions on the final performance~\cite{kendall2017geometric}. PoseCNN~\cite{YuXiang17_PoseCNN} aims to disentangle the pose estimation into several sub-tasks, i.e. regressing object masks, estimating the translation of the object centroids, and regressing quaternions for rotation. Alternatively, the SSD6D detector~\cite{kehlSSD6DMakingRGBbased2017b} relies on a discrete viewpoint classification rather than direct regression of rotations. To make it possible, all poses are divided into a large number of discrete ones, and each of them is further divided into a smaller number of discrete in-plane rotations. As a result, this significantly reduces the solution space allowing for simpler training of the network. However, rotation discretization and approximation of translation lead to poor pose estimates, which unconditionally need further refinement. 
Despite their seeming simplicity, the direct pose regression methods report significantly lower accuracy when compared to correspondence-based solutions. Therefore, post-processing pose refinement, e.g. ICP, must be performed to overcome the issue. DenseFusion~\cite{wang2019densefusion} was free of some of those problems by using RGBD input directly.  Another notable exception is the CosyPose introduced in~\cite{labbe2020cosypose}, which is the best performing method on several datsets from the BOP challenge~\cite{bopchallenge}. The paper is a conceptual extension of PoseCNN and DeepIM~\cite{li2018deepim}, which achieves exceptionally good results by using better deep learning backbones and a better pose parameterization. Additionally, CosyPose~\cite{labbe2020cosypose} introduced a multi-view refinement method which boosted the results even further.

\begin{figure*}
    \centering
    \includegraphics[width=1\linewidth]{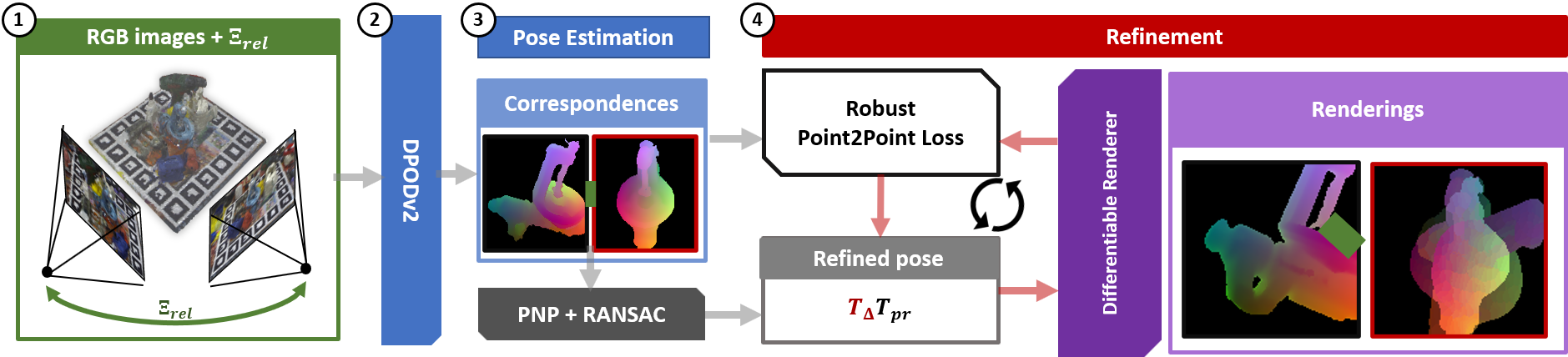}
    \caption{\textbf{Multi-view pose optimization:} The algorithm takes the output of DPODv2 from several views with known relative camera transformations as input. An initial pose hypothesis is iteratively refined until it converges to a pose, which is consistent with predicted correspondences in all frames. The proposed loss function penalizes pixel-wise distance between a predicted correspondence and a correspondence corresponding to the current pose hypothesis. The loss function is implemented using a differentiable renderer.}
    \label{fig:pipeline_refiner}
\end{figure*}

\subsection{Template-Based Methods}

In template-based methods, a template represents a global descriptor capturing the object's appearance under a certain view angle. A set of such templates is extracted to cover all possible object's views. At the test time, the entire template set is compared to the input image at all possible locations using a predefined matching score and the best candidates are pruned to get the final result.

Arguably the most popular template-based methods for 6 DoF pose estimation is LineMOD -- a instance template method \cite{hinterstoisserMultimodalTemplatesRealtime2011b,hinterstoisser2012gradient} relying on a combination of the RGB gradients and surface normals computed from depth images. Thanks to a very efficient implementation, the method is capable of covering an impressive number of angle and scale combinations and still estimate the pose in real time. A follow-up work~\cite{hinterstoisser2012model} demonstrated how to extract templates from synthetic 3D models and then apply them to real data. This was of great value to industrial applications since all possible views and scales for a given object can be generated at no cost. The further follow-up works aimed to improve the scalability issue related to the used number of templates allowing for significant speed-ups~\cite{konishi2016fast, rios-cabreradiscriminativelytrainedtemplates2013,kehlhashmod,hodavn2015detection}.

With the advent of deep learning, more and more methods started to rely on CNNs to learn descriptors directly from data as opposed to using hand-crafted templates. The method introduced in~\cite{wohlhart2015learning} mapped image data directly to the similarity-preserving descriptor space. To learn the mapping, the authors utilized a triplet network and a cost function defined such that the squared Euclidean distance between similar objects was minimized and for dissimilar objects the hinge loss was applied forcing them to be pulled apart using a margin term. The matching is then done by applying efficient and scalable nearest neighbor search methods on the descriptor space to retrieve the closest neighbors. The follow-up works ~\cite{zakharov2017}, ~\cite{bui2018regression} extend the method by introducing an additional rotational degree of freedom (making it 3 DoF), improving the descriptor space by introducing a modified loss function, and a multi-task extension combining manifold learning with direct pose regression leveraging the advantages of both. Another line of works~\cite{zakharov2018keep,planche2019seeing} strive to improve the domain generalization capabilities of the method making it possible to train it from synthetic data while performing on par and better than real methods by using the inverse domain adaptation technique. Building on the success of the above methods, the AAE~\cite{sundermeyerImplicit3DOrientation2018} method proposed a full 6 DoF pose estimation pipeline based on learned descriptors by using already computed SSD detections as input.

Despite their popularity, the template based methods have a number of drawbacks. The first one is related to a limited discrete number of view templates. As a result, the pose estimates  are often very imprecise and additional post-processing refinement methods are required, e.g., Iterative Closest Point (ICP)~\cite{besl1992method}. Using a large number of templates can improve the accuracy, but in return leads to a slow matching during the test phase. Moreover, template-based methods are often very sensitive to occlusions and clutter since they rely on global descriptors. However, despite the drawbacks, these methods are still very popular due to their simplicity and proven efficiency.

\subsection{Correspondence-Based Methods}

A popular alternative to template-based methods are methods utilizing correspondences between the input image and the 3D object model. Their power lies in a reduced solution space and explicitly introduced geometric constraints. This results in more precise pose estimates and the increased robustness to occlusions and clutter.

\setlength{\tabcolsep}{13pt}
\begin{table*}[t]
\centering
  \centering
  \caption{Data modalities and types of train data used in the experiments.\label{tab:data_modalities}}
  
  \resizebox{1\textwidth}{!}{%
  
     \begin{tabular}{l|cc|cccc}
          & \multicolumn{2}{c|}{YOLO} & \multicolumn{4}{c}{CENet} \\
    \midrule
    %TODO ontent from here

    % \midrule
    Dataset/Modality & Real RGB & Synt. RGB & Real RGB & Synt. RGB & Real depth & Synt. depth \\
    \midrule
    Linemod & Yes   & Yes   & Yes   & Yes   & Yes   & Yes \\
    Occlusion & Yes   & Yes   & Yes   & Yes   & Yes   & Yes \\
    TLESS & Yes   & Yes   & Yes   & Yes   & No    & Yes \\
    HomebrewdDB & No    & Yes   & No    & Yes   & No    & Yes \\%

    %TODO content till here
    \end{tabular}%
 }
    
\end{table*}%

The first family of methods rely on prediction of a pre-defined set of keypoints whose 3D locations on the object model are known, followed by the PnP algorithm to compute the full 6 DoF pose. Rad et al. presented a holistic three-stage approach called BB8~\cite{radBB8ScalableAccurate2017b}. In the first two stages, a coarse-to-fine segmentation is performed, the result of which is then fed to the third network trained to output projections of the object's bounding box points. Knowing 2D-3D correspondences, a 6 DoF pose can be estimated using a PnP (Perspective-n-Point) solver. The main disadvantage of this pipeline is its multi-stage nature, resulting in slow run times. Building on the popular 2D detector YOLO~\cite{redmonyouonlylook2015} and the BB8 ideas, YOLO6D~\cite{tekinRealTimeSeamlessSingle2017} proposed a novel deep learning architecture capable of efficient and precise object detection and pose estimation without refinement. As is the case with BB8, the key feature here is to perform the regression of the reprojected bounding box corners in the image. Once correspondences are predicted, the pose can be estimated using a EPnP solver~\cite{lepetit2009epnp}, similarly to BB8.

Another alternative is to use dense pixel-wise 2D-3D or 3D-3D correspondences between image pixels and the object model. The assumption is that a larger number of correspondences will mitigate the problem of their inaccuracies and will result in more precise poses. Moreover, it allows for a significantly better treatment of occlusions. Arguably the first method to formulate pose estimation as dense 2D-3D correspondence prediction was ~\cite{brachmann2014learning}. The key idea was later revisited in several publications by switching from random forests to deep learning.
 Some of the most recent representatives include iPose~\cite{jafari2018ipose}, PVNet~\cite{SPeng18_PVNet}, DPOD~\cite{zakharov2019dpod}, Pix2Pose~\cite{park2019pix2pose}, CDPN~\cite{li2019cdpn}, and SDFlabel~\cite{zakharov2020autolabeling}.
iPose~\cite{jafari2018ipose} operates in 3 stages: segmentation, 3D coordinate regression and pose estimation. In contrast, DPOD~\cite{zakharov2019dpod} and Pix2Pose~\cite{park2019pix2pose} unify the first two stages into the end-to-end network. Moreover, DPOD~\cite{zakharov2019dpod} classifies correspondences instead of regressing them that turned out to be a much easier task for the network, resulting in less erroneous correspondences. CDPN~\cite{li2019cdpn} disentangles 6 DoF pose estimation by using 3D correspondences for rotation estimation and regressing translation directly from the image. EPOS~\cite{hodan2020epos} proposed a different dense corresponding parameterization. Instead of a simple prediction of 3D locations of each 2D point, the network predicts two values. First, it classifies a pixel according to which object segment it belongs to. Then, a precise coordinate within the segment is regressed. This parameterization allows for easier handling of object symmetries. Combined with a robust and efficient variant of the PnP-RANSAC algorithm, the method achieves excellent results on the TLESS dataset.  SDFlabel~\cite{zakharov2020autolabeling} uses 3D correspondences to estimate the pose and additionally estimates the shape of the object. NOCS~\cite{wang2019normalized} showed that dense correspondences prediction can generalize to objects not seeing during training if the network is trained and tested on the same class of objects. PVNet~\cite{SPeng18_PVNet} takes a different approach and designs a network which for every pixel in the image regresses an offset to the predefined keypoints located on the object itself. The ideas on PVNet were further developed in several works. PVNet3D~\cite{he2020pvn3d} takes both RGB and and depth images as input. After independent feature extraction from each data modality, the features are merged and Hough voting is performed to estimate locations of the pre-defined keypoints in 3D space. Then, a least-square algorithm is used to estimate the pose. Alternatively, HybridPose~\cite{song2020hybridpose} extends the initial PVNet approach by also  predicting edges and axes of symmetries. They are used together with the predicted keypoints to estimate the pose with the modified PnP algorithm.

In contrast to the aforementioned dense correspondence methods, we show that DPODv2 can be used on data of various modalities, i.e. RGB and RGBD, and can be trained both on real and on synthetic data. Moreover we also introduce 
a generic correspondence-based pose refinement approach, which does not require any retraining. Finally, we provide a thorough evaluation of all the possible approaches to measure how those choices affect quality of dense correspondences and, subsequently, the quality of 6 DoF poses.

\subsection{Monocular Pose Refinement}

Numerous methods for monocular pose refinement have been introduced over the years. One of the most fundamental and widely used methods is Iterative Closest Point~\cite{besl1992method} (ICP). It works by iteratively aligning an observed point cloud with object's vertices starting from an initial pose hypothesis. %A common fundamental limitation of ICP and ICP-like algorithms is that they require reliable registered depth information, which is not always readily available. 
If only RGB image are available, the object pose can be refined using edge-based alignment, which in spirit resembles ICP. %The core principle is to iteratively minimize the discrepancy between the edges of the object rendered in the initial predicted hypothesis and the edges in the image. 
This approach is very sensitive to the image quality, clutter, occlusions and to the way in which the edge correspondences are computed. There have also been attempts to replace traditional optimization-based hand-crafted refiners with deep learning~\cite{zakharov2019dpod,manhardt2018deep,li2018deepim,radBB8ScalableAccurate2017b}. They share a common principle of training a network to compare a detected object with a rendering of the object in the initial posy hypothesis. In BB8~\cite{radBB8ScalableAccurate2017b}, a network was trained to predict an offset for predicted keypoints, which will make the pose more precise. DPOD~\cite{zakharov2019dpod}, DeepIM~\cite{li2018deepim} and ~\cite{li2018deepim} design the network to directly predict the pose offset by re-parameterizing the pose disentangling rotation and translation. While excellent pose quality was reported in these papers, the methods are rather inconvenient because they require training for each particular model and the effectiveness of pose refinement depends heavily on the correct initial error prior used during training. Unlike these approaches our approach is an optimization approach and requires no lengthy retraining for each new object.  

\begin{figure*}[!t]
\centering
\begin{subfigure}{.5\textwidth}
  \centering
  \includegraphics[width=.82\linewidth]{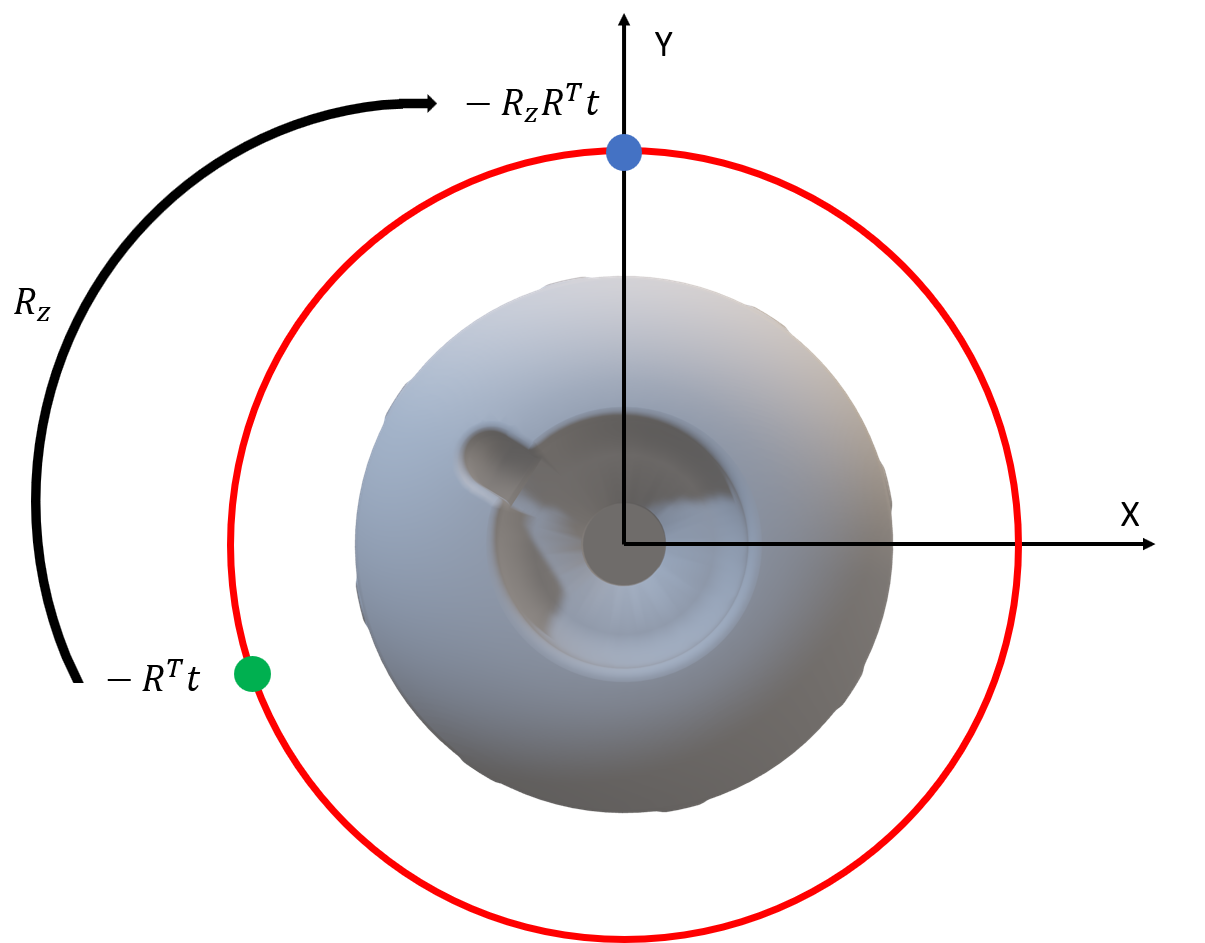}
  \caption{A continuous symmetry around the Z axis \label{fig:symm_z}}
\end{subfigure}%
\begin{subfigure}{.5\textwidth}
  \centering
  \includegraphics[width=.7\linewidth]{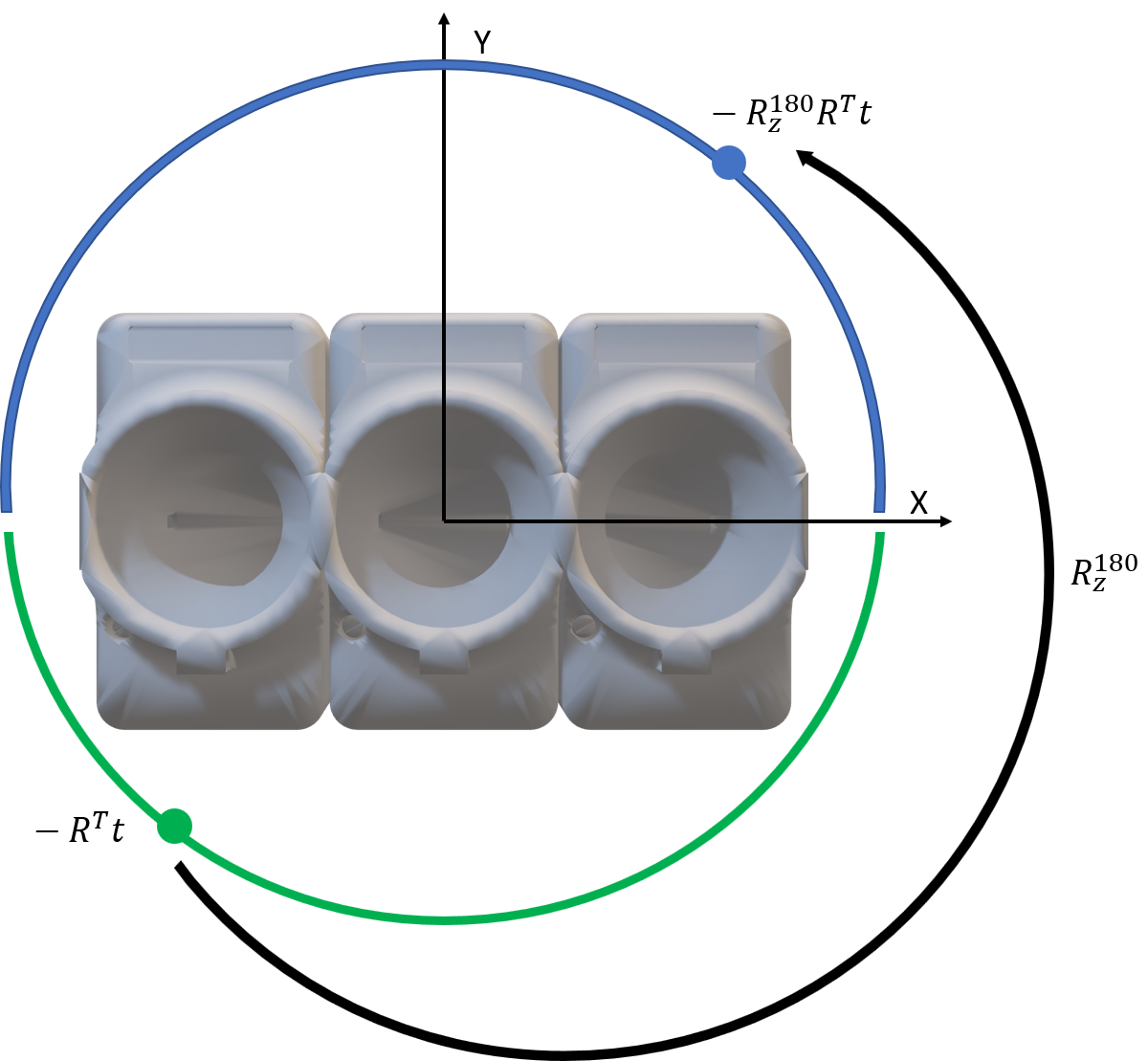}
  \caption{A discrete symmetry around the X axis \label{fig:symm_x}}
\end{subfigure}
\caption{\textbf{Pose recalculation for symmetric objects during training.} Poses are disambiguated during training to produce consistent NOCS maps. In case of a continuous symmetry around Z axis (Example a), a rotation around Z is added to the initial pose to ensure that the camera is always located on the same arch in the object coordinate system.  In case of discrete symmetries (Example b), all symmetric poses are mapped to the one base pose by rotation around the symmetry axis.}
\label{fig:symm}
\vspace{-2em}
\end{figure*}

\subsection{Pose Estimation From Multiple Views}

There have been several attempts to take advantage of using several images to localize the object and estimate the pose. The most prominent examples are~\cite{sock2017multi,erkent2016integration,li2018unified,kaskman2020,labbe2020cosypose,shugurov2021multi}. The potential advantages of multi-view methods are better handling of occlusions, because the object is observed from different viewpoints, and better pose estimation facilitated by geometric multi-view constraints.  However, to the best of our knowledge, the method of~\cite{kaskman2020} is the only one which actually uses several views to detect the object jointly in each frame and estimate a single joint pose.  This is achieved by first obtaining a sparse 3D reconstruction from several views and then object detection and pose estimation in the reconstructed point cloud. Methods in \cite{sock2017multi,erkent2016integration,li2018unified}, on the other hand, use a monocular pose estimation network separately in each frame and use multiple frames with known relative camera transformations only for pose selection or pose refinement. CosyPose~\cite{labbe2020cosypose} follows the same paradigm as the aforementioned methods but does not require known relative camera poses. Instead, it uses a RANSAC scheme to match pose hypotheses from several frames and produce a unified object-level scene reconstruction and approximate relative camera poses. They are jointly optimized by minimizing the multi-view reprojection error.

%% file: method.tex
\section{Methodology}

In this paper, we propose a three-stage object detection and pose estimation pipeline with a potential fourth refinement stage. It builds atop of the existing state of the art results with training on both real and synthetic data of potentially different modalities. The pipeline is visualized in Figure~\ref{fig:pipeline}. The first stage is responsible for 2D object detection from RGB images, which is a standard and well-studied research area. In the second stage, a convolutional neural network, specifically designed for semantic segmentation, takes a detected patch with an object and regresses dense correspondences between input pixels and the 3D object model. Object poses is computed using the predicted correspondences in the third stage. The optional fourth stage, visualized in Figure~\ref{fig:pipeline_refiner}, is responsible for pose refinement, as discussed later. Such a setup, allows for flexible mixing of data modalities and types of training data. Additionally, it simplifies the training procedure, as it breaks the task into smaller independent subtasks. In this section, we provide detailed explanations of each part. Then, we introduce and explain the multi-view refiner which operates atop of dense correspondences predicted by DPODv2.

\subsection{2D Object Detection}

The first step of the pipeline consists of an off-the-shelf 2D object detector. As shown in previous work of~\cite{kehlSSD6DMakingRGBbased2017b,sundermeyer2018implicit,sundermeyer2020augmented}, it is easier to achieve good 2D detection recall if it is done by conventional object detection approaches. SSD6D, which was trained purelly on synthetic data, relied on fine-tuning a separate confidence threshold per each object class to achieve near-perfect recall. AAE\cite{sundermeyer2018implicit,sundermeyer2020augmented}, on the other hand, trained their 2D detector completely on real data. We used YOLOv3\cite{redmon2018yolov3} in all our experiments, although the pipeline is agnostic to a particular choice of the detector. YOLO was trained either completely on real or completely on synthetic RGB data. Another crucial reason for splitting object detection and correspondence estimation into two disjoint steps is scalability. Most of the approaches, mentioned in the related work, train a separate detector per each object to improve the results. While it is achievable on smaller datasets, such  as  Linemod~\cite{hinterstoisser2012model}, it is more challenging on more sophisticated datasets \cite{kaskman2019homebreweddb,hodan2017tless} and close to impossible in real life.

\subsection{Pose Parameterization With Dense Correspondences}

Instead of direct pose regression, which is still challenging and ill-posed for deep learning, we advocate the use of dense per-pixel correspondences between the image and the object model. With the correspondences at hand, the pose can be estimated with a number of well-studied methods depending on the data modality of choice~\cite{hartley2003multiple}. While in the original DPOD paper, the authors relied on well-known 2-dimensional UV maps from computer graphics, we opted for the three-dimensional Normalized Object Coordinates Space (NOCS)~\cite{wang2019normalized} maps. The main reason is its simplicity and the lack of need for manual adjustment of the UV maps. Each dimension of NOCS corresponds to a dimension of the object scaled uniformly to fit in $[0, 1]$ range. This parameterization allows for trivial conversion between the object coordinate system and the NOCS coordinate system, which is more suitable for regression with deep learning due to its constrained nature. 

\begin{figure*}[!t]
\centering

\begin{subfigure}{.4\textwidth}
  \centering
  
  \hfill
  \includegraphics[width=.45\linewidth]{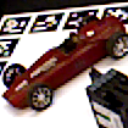}
  \hfill
  \includegraphics[width=.45\linewidth]{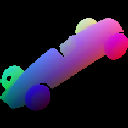}
  \hfill
  
  \hphantom{}
   
 \hfill
 \subfloat[Input patch]{
  \includegraphics[width=.45\linewidth]{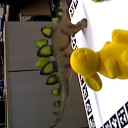}
  }
  \hfill
  \subfloat[GT NOCS]{
  \includegraphics[width=.45\linewidth]{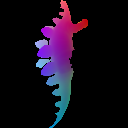}
  }
  \hfill

\end{subfigure}%
\begin{subfigure}{.4\textwidth}
  \centering

  \hfill
  \includegraphics[width=.45\linewidth]{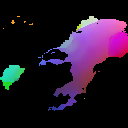}
  \hfill
  \includegraphics[width=.45\linewidth]{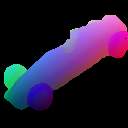}
  \hfill
  
  \hphantom{}
  
  \hfill
  \subfloat[Prediction from Depth]{
   \includegraphics[width=.45\linewidth]{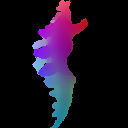}
   }
  \hfill
  \subfloat[Prediction from RGB]{
  \includegraphics[width=.45\linewidth]{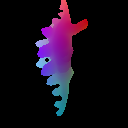}
  }
  \hfill
  
\end{subfigure}

\caption{Visual comparison of predicted segmentation maps of the visible object parts and color-coded NOCS for the same patch depending on whether RGB or Depth CENet is used.\label{fig:segmentation_comparison}}
\vspace{-2em}

\end{figure*}

Let us formally define a model as a set of its vertices: $\mathcal{M} \coloneqq \{v \in \mathbb{R}^3\}$. Operators that compute minimal and maximal coordinate along the vertex dimension $i$ of all $v \in \mathcal{M}$ are defined as 

\begin{equation}
\begin{aligned}[c]
min_i(\mathcal{M}) \coloneqq \minunder_{v \in \mathcal{M}}v_i,
\end{aligned}
\qquad\qquad
\begin{aligned}[c]
max_i(\mathcal{M}) \coloneqq \maxunder_{v \in \mathcal{M}}v_i
\end{aligned}
\label{eq:min_max_operator}
\end{equation}

Then, for any model's vertex $v$, the NOCS projection operator is defined w.r.t. the model as 
\begin{equation}
    \pi_{\mathcal{M}}(v) \coloneqq \left\{ \frac{v_d - min_d(\mathcal{M})}{max_d(\mathcal{M})-min_d(\mathcal{M})}\right\}_{d \in \left\{x, y, z\right\}}
    \label{ea:nocs}
\end{equation}
The inverse of the transformation is denoted by $\pi_{\mathcal{M}}^{-1}$. We use the standard $SE(3)$ definition of 6 DoF object pose~\cite{hartley2003multiple}. A model with the given parameterization can easily be rendered to produce pixel-wise ground truth 2D-3D correspondences. Thus, it allows for instant establishment of 2D-3D and 3D-3D correspondence if RGB or depth data is used respectively.

\subsection{CENet: Correspondence Estimation Network}

The architecture of the Correspondence Estimation Network builds on the architecture of DPOD~\cite{zakharov2019dpod}. The network is an encoder-decoder convolutional neural network with skip connections resembling UNet~\cite{ronneberger2015u}. The network accepts the input of size $\mathcal{I} \in  \mathbb{R}^{128\times128\times D}$, where $D$ stands for the number of input channels and depends on the data modality. The network has a common encoder and four separate decoder heads. One head is responsible for predicting binary per-pixel segmentation mask $\tilde{\mathcal{S}} \in \mathbb{R}^{128\times128\times2}$, while the other three output tensors, each of which $\tilde{\mathcal{C}}_{d} \in \mathbb{R}^{128\times128\times 256}$ after softmax corresponds to pixel-wise probabilities of discretized NOCS coordinates. The encoder is based on the 34-layer ResNet~\cite{heDeepResidualLearning2016} architecture, which proved to be both sufficiently effective and fast. The decoders upsample the computed features up to its original size using a stack of bilinear interpolations followed by convolutional layers. We used biliinear upsampling instead of upconvolutions to make inference more rapid and memory-efficient. However, in principle any other architecture for semantic segmentation could be used. 

Formulating NOCS coordinate regression problem as discrete classification problem proved to be useful for much faster convergence and for the superior quality of correspondences as was also confirmed in previous work~\cite{zakharov2019dpod,wang2019normalized}. Initial experiments on direct coordinate regression showed very poor results in terms of correspondence quality. The main reason for the problem was the infinite continuous solution space, i.e., $[0; 1]^3$, where $3$ is the number of dimensions and $[0, 1]$ is the normalized coordinate range of a 3D model. Classification of the discretized 2D correspondences allowed for a large boost of the output quality by dramatically decreasing the output space, which is now $256^3$ with $256$ being the size of a discretized NOCS dimension.

As a result, the network is trained by optimizing two classification losses. Segmentation loss is defined as a per-pixel binary Dice loss~\cite{milletari2016v} $\mathcal{L}_{DICE}$, which helps overcome the imbalanced number of foreground and background pixels and NOCS classification cross entropy loss defined only for foreground pixels (pixels occupied with non-occluded objects of interest) separately per each NOCS dimension $d$:

\begin{equation}
\mathcal{L}_{NOCS}^d =  \sum_{i=1}^{128}\sum_{j=1}^{128} \mathbbm{1}[\mathcal{I}_{i,j} \text{ is foreground}] \mathcal{L}_{cls}(\tilde{\mathcal{C}}_{d,i,j}, \mathcal{C}_{d,i,j}^{gt})
\end{equation}

Which results in the total per-image loss:
\begin{equation}
\mathcal{L} = \alpha \mathcal{L}_{DICE} + \sum_{d=1}^3 \mathcal{L}_{NOCS}^d
\end{equation}
where $\alpha$ is a weight factors set to $5$ in all our experiments.

\subsection{Inference with The Correspondence Estimation Network}

Given YOLO predictions, we explored three different ways to first get the dense correspondences and then utilize them for pose estimation. The first approach follows the conventional DPOD paradigm of running an RGB correspondence network followed by the PnP with RANSAC. Alternatively, if depth maps are available, it is possible to project the predicted correspondences from 2D into 3D and then use the Kabsch algorithm with RANSAC to predict the pose from 3D-3D correspondences. Depth maps can also directly be used instead of RGB for inferring segmentation masks and 3D-3D correspondences. In this case, Kabsch+RANSAC is again used for pose estimation.

\subsection{Multi-View Refinement With Differentiable Renderer}

The overall pipeline of the proposed refiner is provided in Figure~\ref{fig:pipeline_refiner}. The key idea of the refiner is to predict NOCS correspondences and poses separately for each frame with DPODv2 as described above. Then, if several views with known relative camera positions are available, the predicted poses can be refined to produce a pose which is plausible given correspondences in all frames and the geometric constraints imposed by the relative camera poses. This is achieved by defining a loss function which penalizes pixel-wise difference between predicted NOCS coordinates with CENet, and NOCS coordinates corresponding to the object rendered in the predicted camera poses. The loss function is implemented using the differentiable renderer and minimized iteratively by computing a pose update $\mathbf{T}_\Delta \in SE(3)$ using gradient descent at each step.

We rely on the assumption of the availability of relative camera transformations. There are multiple ways how they can be estimated. Camera transformations can be obtained by placing the object on the markerboard and either using an actual multi-camera system or using a single camera but moving the markerboard. In the scenario of robotic grasping, a camera can be mounted on the robotic arm to enable observation of the object from several viewpoints. Alternatively, SLAM methods could be used to estimate the transformations. While relative transformations are assumed to be known, the object poses still remain unknown and cannot be inferred directly from the relative camera transformations.

The differentiable renderer is used for a differentiable definition of the following functions: binary foreground/background rendering $S: \mathcal{M} \times SE(3) \rightarrow \{0, 1\}^{W \times H}$ and NOCS rendering $C: \mathcal{M} \times SE(3) \rightarrow [0, 1]^{W \times H \times 3}$. We will omit their dependence on the CAD model $\mathcal{M}$ to keep the notation concise. For each set of images used for multi-view refinement, one of the images is taken as the reference frame, and then for each $f$-th image,  its relative pose $\mathbf{\Xi}_{ref\rightarrow f} \in SE(3)$ is recomputed w.r.t. to the reference frame. Initial pose hypotheses $\mathcal{T} \coloneqq \left\{ \mathbf{T}_1, ..., \mathbf{T}_N \right\}$ are estimated separately for each frame with PnP+RANSAC. The pose of the object in the coordinate system of the reference frame is later denoted by $\mathbf{T}_{pr}$. $\mathbf{T}_{s} \in SE(3)$ is a symmetry transformation associated with each object, which must be adjusted to produce consistent NOCS maps for symmetric objects. Computation of $\mathbf{T}_{s}$, described in the data preparation section, is closed-form and quick but has to be performed for each object in each frame after each iteration.

For a pixel $p$ in the $f$-th frame, the loss function is defined as follows:
\begin{equation}
    \mathcal{L}_{f, p} \coloneqq \rho\left(\pi^{-1}\left(\tilde{\mathbf{C}}_{f,p}\right), \pi^{-1}\left(C\left(\mathbf{\Xi}_{ref\rightarrow f} \cdot \mathbf{T}_{\Delta} \cdot \mathbf{T}_{pr} \cdot \mathbf{T}_{s}\right)_p\right)\right)
\end{equation}
where $\rho$ is an arbitrary distance function. In a nutshell, $\tilde{\mathbf{C}}_{f,p}$ is a predicted NOCS correspondence, while $C\left(\mathbf{\Xi}_{ref\rightarrow f} \cdot \mathbf{T}_{\Delta} \cdot \mathbf{T}_{pr} \cdot \mathbf{T}_{s}\right)_p$ projects the model in the pose $ \mathbf{T}_{\Delta} \cdot \mathbf{T}_{pr}$ onto the corresponding frame and computes a NOCS coordinate for that pose. The inverse transformation allows us to relate the pixel-wise discrepancy to the actual error in the 3D coordinate system of the model.

On the level of the full set of images used for refinement, the objective of the refinement is defined as:
\begin{equation}
\resizebox{1\linewidth}{!}{
    $\mathbf{T}_\Delta^{*} = \argmin_{\mathbf{T}_\Delta \in \mathcal{SE}(3)} \sum_{f=1}^{N} \sum_{p \in \mathcal{I}} \left[\tilde{\mathbf{S}}_{f, p} \cdot S\left(\mathbf{\Xi}_{ref\rightarrow f} \cdot \mathbf{T}_{\Delta} \cdot \mathbf{T}_{pr} \cdot \mathbf{T}_{s}\right)_p\right] \cdot \mathcal{L}_{f, p}$
    }
\end{equation}
Here, $\left[\tilde{\mathbf{S}}_{f, p} \cdot S\left(\mathbf{\Xi}_{ref\rightarrow f} \cdot \mathbf{T}_{\Delta} \cdot \mathbf{T}_{pr}\right)_p\right]$ serves as an indicator function which shows whether or not the pixel is foreground in both the rendered and the predicted NOCS maps.

Choice of the reference frame might have a significant effect on the effectiveness of pose refinement, especially in case of occlusions which cause bad initial poses. We aim at choosing a reference frame in such a way that its pose has the lowest error when re-projected to other frames. To deal with degenerate cases when there is no overlap or very small overlap between the projected model and the predicted foreground segmentations, the loss is scaled by their intersection over union as follows:
\begin{equation}
%\resizebox{1\linewidth}{!}{
   \argmin_{ref \in \left[1, .., N\right]} \frac{1}{K} \sum_{f=1}^N 
    \frac{
    \sum_{p \in \mathcal{I}} \left[\tilde{\mathbf{S}}_{f, p} \cdot S\left(\mathbf{\Xi}_{ref\rightarrow f} \cdot \mathbf{T}_{ref} \cdot \mathbf{T}_{s}\right)_p\right] \cdot \mathcal{L}_{f, p}
    }
    {
    IOU\left(\tilde{\mathbf{S}}_{f}, S\left(\mathbf{\Xi}_{ref\rightarrow f} \cdot \mathbf{T}_{ref} \cdot \mathbf{T}_{s}\right)\right)
    }
%    }
\end{equation}
where $K$ stands for the number of frames with non-zero loss. Additionally, $\argmin$ ignores zero values, as they correspond to degenerate poses that are not re-projected correctly onto other frames.

\subsection{Implementation Details}

Our pipeline is implemented using the Pytorch~\cite{paszke2017automatic} deep learning framework. All the experiments were conducted on an Intel Core i9-9900K CPU 3.60GHz with NVIDIA Geforce RTX 2080 TI GPU.
YOLO was trained with the ADAM optimizer with a constant learning rate of $1 \times 10^{-3}$ and weight decay of $3 \times 10^{-5}$ for 100 epochs. Correspondence estimation network was trained for 240 epochs with the learning rate of $5 \times 10^{-5}$ which was decreased after 40th, 120th and 200th epochs. We used EPnP~\cite{lepetit2009epnp}+RANSAC implementation from OpenCV~\cite{opencv_library}  and point-to-plane ICP from Open3D~\cite{Zhou2018}. For Kabsch+RANSAC we used our own minimalistic implementation.

The correspondence inference network uses 128$\times$128 images for training and for testing, while YOLO is trained to predict tight bounding boxes around the objects. Therefore, the shortest side of each bounding box predicted by YOLO is padded to match the longest side, after which the corresponding image patch is resized preserving the aspect ratio.  During training, small random imprecisions are added to each side of the ground truth bounding box to ensure that CENet is robust to noisy YOLO predictions. The ablation studies show the effectiveness of this approach, as the quality of predicted correspondences does not deteriorate much when the predicted bounding boxes are used instead of the ground truth boxes.

When training on synthetic data, the problem of domain adaptation becomes one of the main challenges. Training the network without any prior parameter initialization makes generalization to the real data difficult if possible at all. A simple solution to this problem was proposed in several works, including \cite{hinterstoisser2018pre, manhardt2018deep}, where they freeze the first layers of the network trained on a large dataset of real images, e.g., ImageNet~\cite{ImageNetLargescaleHierarchical} or MS COCO~\cite{linMicrosoftCOCOCommon2014a}, for the object classification task.  The common observation of the authors is that these layers, learning low-level features, very quickly overfit to the perfect object renderings. We found out that this setup is only partially needed for our networks. We froze first 20 layers when we trained YOLO. The weights were initialized from the YOLO trained on the MS COCO dataset~\cite{linMicrosoftCOCOCommon2014a}. In the correspondence estimation network, the weights were initialized from the Resnet pretrained on ImageNet~\cite{ImageNetLargescaleHierarchical}  and then were all modified during training.  When the correspondence network was trained on depth maps, the first layer was instead initialized randomly and then all layers were modified during training. We have found out that replacing BatchNorm~\cite{ioffe2015batch} with InstanceNorm\cite{ulyanov2016instance} helps overcome overfitting and allows for reliable and stable training in this case. 

We used the Soft Rasterizer renderer~\cite{liu2019soft} in all experiments with the multi-view refiner. The general robust function~\cite{barron2019general} was used as a distance measure $\rho$ in the loss function to effectively deal with outliers. Continuous rotation parameterization from~\cite{zhou2019continuity} was used to achieve faster and more stable convergence during the optimization procedure in comparison to quaternions and Euler angles.

%% file: data_preparation.tex
\section{Data Preparation}

When it comes to data preparation for the tasks of object detection and pose estimation, two distinctive paradigms are observed. The most popular and conventional one follows the standard way of machine learning of collecting a large database of labeled data. Unfortunately, pose estimation requires more involved labels, which cannot be produced manually, in contrast to the standard classification or segmentation labels. Pose labeling requires sophisticated multi-step procedures, such as in~\cite{kaskman2019homebreweddb,hodan2017tless}. In spite of those challenges, real label train data remain the predominant choice of data especially because it allows to achieve excellent performance on academic benchmarks. 
In the deep learning on point clouds, real train data are also predominant.

On the other hand, it is possible to leverage advances in computer graphics and to rely solely on synthetically rendered train data if object CAD models are available. The advantages of this include virtually unlimited number of images and poses with no labeling effort.  However, the domain gap between the real and the rendered images comes into play to hinder the detection rate and pose accuracy. In the past, the predominant way to train a deep neural network on synthetic data was to render the object on random backgrounds, as was done in DPOD~\cite{zakharov2019dpod}, SSD6D~\cite{kehlDeepLearningLocal2016a} and in \cite{hinterstoisser2018pre}. On the other hand, complete photo-realistic scene rendering is becoming more popular~\cite{denningerblenderproc}.

\subsection{RGB Data Preparation}

\textbf{Real Data Preparation.} Preparation of real train data is rather straightforward. Given an image, CAD models, and associated ground truth poses, it is enough to render present objects in ground truth poses. Binary segmentation masks and per-pixel NOCS coordinates are obtained by rendering the models in the ground truth poses.  
Following the paradigm of the previous works\cite{tekinRealTimeSeamlessSingle2017,zakharov2019dpod} object patches corresponding to the visible object parts were extracted and inpainted in random backgrounds sampled from MSCOCO~\cite{linMicrosoftCOCOCommon2014a}. This ensures that the network learns object-specific features rather than overfits to backgrounds, leading to better generalization to new unseen scenes. For training the detector, a random number of objects were added to each background. For the correspondence estimation network, only one object is in-painted. The object patch is resized randomly to simulate different distances from camera. Additionally, a random in-plane rotation is performed. Then, we use standard color augmentations used by the backbone networks. When training the correspondence network, random occlusions also have to be added to make the network more robust. Realistically looking occlusions are approximated by sampling a random patch, corresponding to a different object, which is then in-painted randomly atop of the object of interest. Occlusions are added at random, but it is made sure of that the target object is still visible in the patch.

\textbf{Synthetic Data Preparation.} We relied on the PBR images provided by the BOP challenge organizers for the preparation of syntetic training data. Those images were produced by simulation using BlenderProc~\cite{denninger2019blenderproc}. No additional transformations were performed on those images apart from random photometric augmentations, such as brightness augmentation. Usage of the provided PBR images also ensured more fairer comparison of our methods to the other methods on the BOP challenge datasets by making sure that the competing methods were trained on exactly the same data.

\subsection{Depth Data Preparation}
In general, training deep learning networks on depth is harder than on RGB images due to holes in the depth and a much larger range of values. To get rid of empty depth values, we fill them in using nearest neighbor interpolation. To alleviate the range problem, depth maps are parameterized similar to PointNet++\cite{qi2017pointnet++} by replacing a depth value in each pixel with a distance ti to the mean depth in the neighborhood of size $T$:
\begin{equation}
   \mathcal{I}_{i, j} = \mathcal{I}_{i, j}^{raw} - \frac{1}{T^2}\sum_{x=-T}^{T}\sum_{y=-T}^{T} \mathcal{I}_{i-x, j-y}^{raw}
\end{equation}
for $i,j \in \mathbb{R}^{128}$ and $\mathcal{I}^{raw}$ being the original depth image with already interpolated missing values. This depth parameterization ensured simpler and more stable training. We set the neighborhood radius $T$ to 5 in our experiments.

In train time, depth maps undergo several random augmentations. First, random holes of random size are added to the depth map and then interpolated accordingly. Then, random Perlin noise~\cite{perlin1985image} is added to the background, that results in smooth random backgrounds, which is especially important for synthetically generated data. Foreground depth values are augmented with random Gaussian noise.
Synthetic train data was again taken from the synthetic PBR images.

\subsection{Handling Object Symmetries}

While there is a large body of theoretical work on dealing with ambiguities induced by object symmetries, for instance~\cite{pitteri2019object,bui2020eccv,manhardt2019explaining}, we followed a more conventional and hands-on approach. Similarly to the DPOD and SSD6D papers, we relied on training on consistent and unambiguous ground truth NOCS, rather than inferring multiple pose hypotheses as in~\cite{pitteri2019object,manhardt2019explaining}. While DPOD and SSD6D were trained only on images of objects taken from non-ambiguous poses, we used all the images and then disambiguated the poses in order to render consistent NOCS coordinates as visualized in Figure~\ref{fig:symm}. There are two main symmetry types in the used datasets: a continuous rotation symmetry around an axis, when every rotation around the axis results in an identically looking object, and discrete symmetry around an axis.

For objects, which are fully symmetric around an axis, NOCS maps are rendered from a consistent camera location in the object coordinate system as visualized in Figure~\ref{fig:symm_z}. Typically, the object is symmetric around its Z axis. Given an object pose $R \in SO(3), t \in \mathbb{R}^3$, which transforms from the model coordinate system into the camera coordinate system, the camera location in the model coordinate system is given by $-R^\top t$. To produce consistent NOCS maps, we rotate the object around Z axis axis in such a way, that a camera location always lies in the YZ plane of the model system, i.e. $R_z \in SO(3)$ such that the X coordinate of $-R_zR^\top t$ is 0. This makes the object always visible from the same position. $R_z$ can easily be computed in a closed form by computing an angle between vectors $\overrightarrow{\left(\left(-R^\top t\right)_x, \left(-R^\top t\right)_y, 0\right)}$ and $\overrightarrow{\left(0, 1, 0\right)^\top}$. The final disambiguated object pose is defined by $RR_z^\top$ and $t$.

Handling objects with discrete symmetries in respect to the plane is rather straightforward. Assuming for the illustration purposes that rotating the object around Z axis by 180 degrees (denoted by $R_z^{180}$) produces an identically looking object as in Figure~\ref{fig:symm_x}. One of the symmetry sides is taken as the base region of the symmetry, denoted by blue in the figure. Then, a camera location is computed and if the camera lies on the opposite side of the object, the rotation is updated as $RR_z^{180}$. Basically, it ensures that the object is always visible from the same side. Even though~\cite{pitteri2019object} noted that this transformation might be unfavourable for neural networks due to its non-continuity, it works reliably in practice.

%% file: experiments.tex
\section{Experiments}

\subsection{Datasets}

Over the past decade, numerous datasets for object detection and 6 DoF pose estimation have been proposed~\cite{hinterstoisser2012model,brachmann2014learning,hodan2017tless,xiang2017posecnn,drost2017introducing,kaskman2019homebreweddb,rennie2016dataset,doumanoglou2016recovering,tejani2014latent}. However, in this work we focus on the following datasets: Linemod~\cite{hinterstoisser2012model}, Occlusion~\cite{brachmann2014learning} (LMO), TLESS~\cite{hodan2017tless} and Homebrewed~\cite{kaskman2019homebreweddb} (HBD).

\setlength{\tabcolsep}{4pt}
\begin{table*}[!t]
  \centering
  \caption{\textbf{Pose estimation performance on Linemod on RGB images of methods trained on synthetic data}: The table reports the percentages of correctly estimated poses w.r.t. the ADD score. Our approach sets the new state of the art both among the methods trained on real data and the methods trained on synthetic data. Our refiner outperforms other RGB refiners. Run times are provided as they are reported in the original papers using non-identical hardware.\label{tab:lm_rgb_synt}}
  
      \begin{tabular}{c|cccc|cccc}
    \textbf{Train Data} & \multicolumn{8}{c}{\textbf{Synthetic}} \\
    \textbf{Method} & SSD6D~\cite{kehlSSD6DMakingRGBbased2017b} & AAE~\cite{sundermeyer2020augmented}   & DPOD~\cite{zakharov2019dpod}  & {Ours} & SSD6D~\cite{kehlSSD6DMakingRGBbased2017b} & DPOD~\cite{zakharov2019dpod}  & {Ours} & {Ours} \\
    \midrule
    \textbf{Refinement} & \multicolumn{4}{c|}{-}        & DL~\cite{manhardt2018deep}    & DL~\cite{zakharov2019dpod}    & {2 calib. views} & {4 calib. views} \\
\cmidrule{2-9}    Ape   & 2.60  & 3.96  & 37.22 & \textbf{62.14} & -     & 55.23 & 98.54 & \textbf{100.00} \\
    Bvise & 15.10 & 20.92 & 66.76 & \textbf{88.39} & -     & 72.69 & \textbf{100.00} & \textbf{100.00} \\
    Cam   & 6.10  & 30.47 & 24.22 & \textbf{92.51} & -     & 34.76 & 99.67 & \textbf{100.00} \\
    Can   & 27.30 & 35.87 & 52.57 & \textbf{96.66} & -     & 83.59 & \textbf{100.00} & \textbf{100.00} \\
    Cat   & 9.30  & 17.90 & 32.36 & \textbf{86.17} & -     & 65.1  & 99.66 & \textbf{100.00} \\
    Driller & 12.00 & 23.99 & 66.60 & \textbf{90.15} & -     & 73.32 & 99.83 & \textbf{100.00} \\
    Duck  & 1.30  & 4.86  & 26.12 & \textbf{54.86} & -     & 50.04 & 99.68 & \textbf{100.00} \\
    Eggbox & 2.80  & 81.01 & 73.35 & \textbf{98.64} & -     & 89.05 & 97.70 & \textbf{99.04} \\
    Glue  & 3.40  & 45.49 & 74.96 & \textbf{95.41} & -     & 84.37 & 97.70 & \textbf{98.03} \\
    Holep. & 3.10  & 17.60 & 24.50 & \textbf{27.08} & -     & 35.35 & 94.01 & \textbf{99.03} \\
    Iron  & 14.60 & 32.03 & 85.02 & \textbf{98.26} & -     & 98.78 & \textbf{100.00} & \textbf{100.00} \\
    Lamp  & 11.40 & 60.47 & 57.26 & \textbf{91.04} & -     & 74.27 & 99.51 & \textbf{100.00} \\
    Phone & 9.70  & 33.79 & 29.08 & \textbf{74.34} & -     & 46.98 & \textbf{100.00} & \textbf{100.00} \\
    \midrule
    \textbf{Mean} & 9.10  & 28.65 & 50.00 & \textbf{81.20} & 34.1  & 66.42 & 98.95 & \textbf{99.70} \\
    \midrule
    \textbf{Time (ms)} & -     & 24    & -     & 32    & -     & -     & 202.00 & 132.00 \\
    \end{tabular}%
    \vspace{-1em}

\end{table*}%

Linemod~\cite{hinterstoisser2012model} is a classical dataset for object pose estimation. It is still an actively used benchmark even though it dates back to the pre-deep learning era. The datasets provides researchers with 13 3D object models. Each object comes with approximately 1200 annotated images. The images exhibit strong background clutter but almost no occlusions. In experiments with real training data, we used exactly the same training/testing split as in BB8~\cite{radBB8ScalableAccurate2017b}. In the experiments with synthetic data, we used the entire datasets as test set, analogously to SSD6D~\cite{kehlSSD6DMakingRGBbased2017b} and DPOD~\cite{zakharov2019dpod}. We report the standard ADD recall~\cite{hinterstoisser2012model} in all experiments, as it is the most widely reported metric for this dataset.

Occlusion~\cite{brachmann2014learning} is an extension of the Linemod dataset, which consists of one Linemod sequence with 8 annotated objects. This dataset is a step forward to more complex datasets with occlusions of various degrees and a varying number of objects. We use the BOP subsequence of the dataset and report the Average Recall (AR) score of the BOP challenge. This allows us to fairly compare our method to all other methods participating in the challenge.

%In spite of their popularity, Linemod and Occlusion do not have clearly separated train, validation and test splits. It leads to the situation when the images from the same sequence of frames are used for all the splits. It means that performance on the benchmarks might be significantly influenced by overfitting to the backgrounds, particular object poses, particular illumination setups and etc. 

The authors of the Homebrewed~\cite{kaskman2019homebreweddb} advocated the importance of training on synthetic data. Therefore, the datasets provides no train images, but has labeled validation set and a test set with hidden labels, as in other major machine learning datasets. Here, we also report the AR score from the BOP challenge. All ablation studies were conducted on the publicly available validation set.

\setlength{\tabcolsep}{4pt}
\begin{table*}[!b]
  \centering
  \caption{\textbf{Pose estimation performance on Linemod on RGB images of methods trained on real data}: The table reports the percentages of correctly estimated poses w.r.t. the ADD score. Our approach sets the new state of the art both among the methods trained on real data and the methods trained on synthetic data. Our refiner outperforms other RGB refiners. Run times are provided as they are reported in the original papers using non-identical hardware.\label{tab:lm_rgb_real}}
  
  \resizebox{1\textwidth}{!}{%
      \begin{tabular}{c|cccccc|ccccc}
    \textbf{Train Data} & \multicolumn{11}{c}{\textbf{Real}} \\
    \textbf{Method} & Pix2Pose~\cite{park2019pix2pose} & DPOD~\cite{zakharov2019dpod}  & PVNet~\cite{SPeng18_PVNet} & CDPN~\cite{li2019cdpn}  & HybridPose~\cite{song2020hybridpose} & {Ours} & BB8~\cite{radBB8ScalableAccurate2017b}   & PoseCNN~\cite{xiang2017posecnn} & DPOD~\cite{zakharov2019dpod}  & {Ours} & {Ours} \\
    \midrule
    \textbf{Refinement} & \multicolumn{6}{c|}{-}                        & DL~\cite{radBB8ScalableAccurate2017b}    & DeepIM~\cite{li2018deepim} & DL~\cite{zakharov2019dpod}    & {2 calib. views} & {4 calib. views} \\
\cmidrule{2-12}    Ape   & 58.10 & 53.28 & 43.62 & 64.38 & 63.1  & 80.09 & 40.40 & 76.95 & 87.73 & 98.66 & \textbf{100.00} \\
    Bvise & 91.00 & 95.34 & \textbf{99.90} & 97.77 & \textbf{99.9} & 99.71 & 91.80 & 97.48 & 98.45 & \textbf{100.00} & \textbf{100.00} \\
    Cam   & 60.90 & 90.36 & 86.86 & 91.67 & 90.4  & \textbf{99.21} & 55.70 & 93.53 & 96.07 & \textbf{100.00} & \textbf{100.00} \\
    Can   & 84.40 & 94.10 & 95.47 & 95.87 & 98.5  & \textbf{99.60} & 64.10 & 96.46 & 99.71 & \textbf{100.00} & \textbf{100.00} \\
    Cat   & 65.00 & 60.38 & 79.34 & 83.83 & 89.4  & \textbf{95.11} & 62.60 & 82.14 & 94.71 & \textbf{100.00} & \textbf{100.00} \\
    Driller & 76.30 & 97.72 & 96.43 & 96.23 & 98.5  & \textbf{98.91} & 74.40 & 94.95 & 98.80 & \textbf{100.00} & \textbf{100.00} \\
    Duck  & 43.80 & 66.01 & 52.58 & 66.76 & 65.0  & \textbf{79.54} & 44.30 & 77.65 & 86.29 & 98.12 & \textbf{100.00} \\
    Eggbox & 96.80 & \textbf{99.72} & 99.15 & \textbf{99.72} & \textbf{100.0} & 99.63 & 57.80 & 97.09 & 99.91 & 99.68 & \textbf{100.00} \\
    Glue  & 79.40 & 93.83 & 95.66 & 99.61 & 98.8  & \textbf{99.81} & 41.20 & \textbf{99.42} & 96.82 & 97.46 & 98.84 \\
    Holep. & 74.80 & 65.83 & 81.92 & 85.82 & \textbf{89.7} & 72.30 & 67.20 & 52.81 & 86.87 & 99.81 & \textbf{100.00} \\
    Iron  & 83.40 & \textbf{99.80} & 98.88 & 97.85 & \textbf{100.0} & 99.49 & 84.70 & 98.26 & \textbf{100.00} & \textbf{100.00} & \textbf{100.00} \\
    Lamp  & 82.00 & 88.11 & 99.33 & 97.89 & \textbf{99.5} & 96.35 & 76.50 & 97.5  & 96.84 & \textbf{100.00} & \textbf{100.00} \\
    Phone & 45.00 & 74.24 & 92.41 & 90.75 & 94.9  & \textbf{96.88} & 54.00 & 87.72 & 94.69 & \textbf{100.00} & \textbf{100.00} \\
    \midrule
    \textbf{Mean} & 72.40 & 82.98 & 86.27 & 89.86 & 91.3  & \textbf{93.59} & 62.70 & 88.61 & 95.15 & 99.52 & \textbf{99.91} \\
    \textbf{Time (ms)} & 100-167 & 36    & 40    & 30    & 1000  & 31    & 330   & -     & -     & 201.00 & 131.00 \\
    \end{tabular}%
    
    }

\end{table*}%

The TLESS dataset~\cite{hodan2017tless} provides researchers with 30 objects and 20 scenes comprising of various number of objects. The dataset includes two kind of object models: precise untextured CAD models and reconstructions of relatively low quality. There is a clear separation between train and test images, with train images not coming from the test domain. The dataset exhibits three main challenges: 1) texture-less objects, which are also similar to each other, 2) heavy occlusions and 3) various object symmetries.

As the proposed approach can be trained both on real and synthetic data and on data of different modalities, we experimented with each option given the availability of this data type. Table~\ref{tab:data_modalities} summarizes which data types were used and whether synthetic or real data was used in each experiment. YOLO was trained only on RGB data as it was proven that in previous works that object detection networks can be relatively easy trained both on real and synthetic data thanks to transfer learning~\cite{hinterstoisser2018pre,kehlSSD6DMakingRGBbased2017b}. The correspondence estimation network was trained either on RGB images or on depth maps. In our experiments with the Homebrewed dataset, we completely relied on synthetically simulated train data as the dataset does not provide real train data. Real train depth data from the TLESS dataset was also not used as it is too simplistic and has no background depth to train the correspondence estimation network. We also experimented with ICP refinement to see how it compares to the proposed multi-view refiner. Exactly the same parameters of ICP were used on all datasets.

In the remaining of the section a following naming convention for out approach is used:
\begin{itemize}
    \item \textit{RGB} - both YOLO and CENet operate on RGB images, the final pose is computed using PnP+RANSAC;
    \item \textit{RGBD} - YOLO operates on RGB images, while CENet takes a depth map as its input. The final pose is computed with Kabsch+RANSAC;
    \item \textit{RGB + D-Kabsch} - YOLO and CENet use RGB as input, but then the predicted correspondences are projected into 3D, and the pose is computed with Kabsch+RANSAC.
\end{itemize}

\begin{table*}[!t]
  \centering
  \caption{\textbf{Pose estimation performance on Linemod on depth and RGBD images}: The table reports the percentages of correctly estimated poses w.r.t. the ADD score. The proposed detector shows state of the art results both if only real or only synthetic train data is used. Run times are provided as they are reported in the original papers using non-identical hardware. \label{tab:lm_rgbd_synt}}
  
  %\resizebox{1\textwidth}{!}{%
% Table generated by Excel2LaTeX from sheet 'LM RGBD'

    \begin{tabular}{c|ccccccccc}
    \textbf{Train Data} & \multicolumn{9}{c}{\textbf{Synthetic}} \\
    \textbf{Modality} & \textbf{RGBD} & \textbf{Depth} & \textbf{Depth} & \textbf{RGBD} & \textbf{RGBD} & \textbf{RGBD} & \textbf{RGBD} & \textbf{RGB + D-Kabsch} & \textbf{RGB + D-Kabsch} \\
    \midrule
    \textbf{Method} & AAE~\cite{sundermeyer2020augmented}   & PPF~\cite{drost2010model}   & PPF++~\cite{hinterstoisser2016going} & {Ours} & {Ours} & SSD6D~\cite{kehlSSD6DMakingRGBbased2017b} & Brachmann et. al.~\cite{brachmann2014learning} & {Ours} & {Ours} \\
    \midrule
    \textbf{Refinement} & ICP   & ICP   & ICP   & -     & {ICP} & ICP   & iterative & {ICP} & \textbf{-} \\
    \midrule
    Ape   & 20.55 & 86.50 & 98.50 & 87.54 & 91.91 & -     & -     & 98.38 & \textbf{98.79} \\
    Bvise & 64.25 & 70.70 & 99.80 & \textbf{99.92} & \textbf{99.92} & -     & -     & \textbf{99.92} & \textbf{99.92} \\
    Cam   & 63.20 & 78.60 & 99.30 & 96.42 & \textbf{97.84} & -     & -     & 99.75 & 99.75 \\
    Can   & 76.09 & 80.20 & 98.70 & 98.16 & 98.66 & -     & -     & \textbf{99.75} & 99.58 \\
    Cat   & 72.01 & 85.40 & 99.90 & 99.41 & 99.83 & -     & -     & \textbf{100.00} & \textbf{100.00} \\
    Driller & 41.58 & 87.30 & 93.40 & 97.55 & 97.64 & -     & -     & \textbf{98.82} & \textbf{98.82} \\
    Duck  & 32.38 & 46.00 & 98.20 & 90.82 & 94.26 & -     & -     & 98.41 & \textbf{98.96} \\
    Eggbox & 98.64 & 97.00 & 98.80 & 98.48 & 98.80 & -     & -     & \textbf{99.52} & \textbf{99.52} \\
    Glue  & 96.39 & 57.20 & 75.40 & 99.26 & 99.34 & -     & -     & \textbf{99.84} & 99.75 \\
    Holep. & 49.88 & 77.40 & 98.10 & 93.45 & 96.12 & -     & -     & \textbf{97.81} & 97.17 \\
    Iron  & 63.11 & 84.90 & 98.30 & 48.35 & 49.21 & -     & -     & \textbf{100.00} & 99.91 \\
    Lamp  & 91.69 & 93.30 & 96.00 & 50.12 & 50.20 & -     & -     & 99.02 & \textbf{99.59} \\
    Phone & 70.96 & 80.70 & \textbf{98.60} & 96.70 & 97.02 & -     & -     & 98.15 & 98.07 \\
    \midrule
    \textbf{Mean} & 64.67 & 78.86 & 96.38 & 88.94 & 90.06 & 90.90 & 98.3  & 99.18 & \textbf{99.22} \\
    \midrule
    \textbf{Time (ms)} & 224   & -     & -     & 49    & 58    & 100   & 545   & 55    & 49 \\
    \end{tabular}%
    
    %}
    
    \vspace{-1.5em}

\end{table*}%

\begin{table*}[!b]
\vspace{-1em}

  \centering
  \caption{\textbf{Pose estimation performance on Linemod on depth and RGBD images}: The table reports the percentages of correctly estimated poses w.r.t. the ADD score. The proposed detector shows state of the art results both if only real or only synthetic train data is used. Run times are provided as they are reported in the original papers using non-identical hardware. \label{tab:lm_rgbd_real}}
  
  \resizebox{0.9\textwidth}{!}{%
    \begin{tabular}{c|cccccccccc}
    \textbf{Train Data} & \multicolumn{10}{c}{\textbf{Real}} \\
    \midrule
    \textbf{Modality} & \textbf{RGBD} & \textbf{RGBD} & \textbf{RGBD} & \textbf{RGBD} & \textbf{RGBD} & \textbf{RGBD} & \textbf{RGBD} & \textbf{RGBD} & \textbf{RGB + D-Kabsch} & \textbf{RGB + D-Kabsch } \\
    \textbf{Method} & PointFusion~\cite{xu2018pointfusion} & DF~\cite{wang2019densefusion}    & DF~\cite{wang2019densefusion}    & Brachmann et. al.~\cite{brachmann2014learning} & {Ours} & Brachmann et. al.~\cite{brachmannUncertaintyDriven6DPose2016a} & {Ours} & PVN3D~\cite{he2020pvn3d} & {Ours} & {Ours} \\
    \midrule
    \textbf{Refinement} & -     & -     & DL~\cite{wang2019densefusion}    & -     & {-} & iterative & {ICP} & -     & {ICP} & {-} \\
    \midrule
    Ape   & 70.40 & 79.50 & 92.30 & -     & 97.34 & -     & 97.62 & 97.30 & 97.72 & \textbf{98.56} \\
    Bvise & 80.70 & 84.20 & 93.20 & -     & 99.90 & -     & \textbf{100.00} & 99.70 & \textbf{100.00} & 99.90 \\
    Cam   & 60.80 & 76.50 & 94.40 & -     & 99.41 & -     & 99.70 & 99.60 & \textbf{99.90} & \textbf{99.90} \\
    Can   & 61.10 & 86.60 & 93.10 & -     & 99.41 & -     & 99.51 & 99.50 & \textbf{100.00} & 99.80 \\
    Cat   & 79.10 & 88.80 & 96.50 & -     & 99.90 & -     & 99.90 & 99.80 & \textbf{100.00} & \textbf{100.00} \\
    Driller & 47.30 & 77.70 & 87.00 & -     & 97.52 & -     & 98.02 & 99.30 & \textbf{99.90} & \textbf{99.90} \\
    Duck  & 63.00 & 76.30 & 92.30 & -     & 94.74 & -     & 97.09 & 98.20 & 98.31 & \textbf{99.25} \\
    Eggbox & 99.90 & 99.90 & 99.80 & -     & 99.63 & -     & 99.72 & \textbf{99.80} & 99.72 & 99.72 \\
    Glue  & 99.30 & 99.40 & 100.00 & -     & 99.71 & -     & 99.71 & \textbf{100.00} & \textbf{100.00} & \textbf{100.00} \\
    Holep. & 71.80 & 79.00 & 92.10 & -     & 99.53 & -     & 99.71 & 99.90 & 99.71 & 99.61 \\
    Iron  & 83.20 & 92.10 & 97.00 & -     & 98.88 & -     & 99.18 & 99.70 & \textbf{100.00} & 99.90 \\
    Lamp  & 62.30 & 92.30 & 95.30 & -     & 99.14 & -     & 99.04 & 99.80 & \textbf{100.00} & 99.90 \\
    Phone & 78.80 & 88.00 & 92.80 & -     & 99.43 & -     & 99.43 & 99.50 & \textbf{100.00} & \textbf{100.00} \\
    \midrule
    \textbf{Mean} & 73.70 & 86.20 & 94.30 & 98.10 & 98.81 & 99.00 & 99.13 & 99.40 & 99.64 & \textbf{99.73} \\
    \midrule
    \textbf{Time (ms)} & -     & -     & -     & 545   & 49    & -     & 57    & -     & 55    & 49 \\
    \end{tabular}%
    
    }

\end{table*}%

\setlength{\tabcolsep}{8pt}
\begin{table*}[!t]
  \centering
  \caption{\textbf{Pose estimation performance comparison on the Occlusion dataset}: Results are reported in terms of the
Average Recall score. The results prove the effectiveness of the proposed approach on all used data modalities. Run times are provided as they are reported in the BOP challenge~\cite{bopchallenge} using non-identical hardware. PPF-based methods, labeled with $^*$ in the Time column, use only CPU. \label{tab:lmo}}

    \resizebox{0.9\textwidth}{!}{%
    
        \begin{tabular}{c|c|c|c|ccc|c|c}
    \textbf{Train data} & \textbf{Data modality} & \textbf{Method} & \textbf{Refinement} & \textbf{VSD} & \textbf{MSSD} & \textbf{MSPD} & \textbf{AR} & \textbf{Time (s)} \\
    \midrule
    \multirow{16}[4]{*}{PBR} & \multirow{8}[2]{*}{RGBD} & CosyPose~\cite{labbe2020cosypose} & ICP   & 	0.567 & 0.748 & 0.826 & 0.714 & 8.289 \\
          &       & Ours + Kabsch & - & 0.565 & 0.748 & 0.788 & 0.700 & 0.334 \\
          &       & Ours + Kabsch & ICP & 0.557 & 0.749 & 0.787 & 0.698 & 0.387 \\
          &       & PVNet~\cite{SPeng18_PVNet} & ICP   & 0.502 & 0.683 & 0.73  & 0.638 & - \\
          &       & CDPN~\cite{li2019cdpn}  & ICP   & 0.469 & 0.689 & 0.731 & 0.630 & 0.506 \\
          &       & Pix2Pose~\cite{park2019pix2pose} & ICP   & 0.473 & 0.631 & 	0.659 & 0.588 & 5.191 \\
          &       & Ours & ICP & 0.472 & 0.621 & 0.654 & 0.582 & 0.398 \\
          &       & Ours & - & 0.422 & 0.58 & 0.621 & 0.541 & 0.325 \\
\cmidrule{2-9}          & \multirow{8}[2]{*}{RGB} & Ours & 4 calibrated views & 0.572 & 0.735 & 0.777 & 0.695 & 2.479 \\
          &       & Ours & 2  calibrated  views & 0.520 & 0.690 & 0.771 & 0.660 & 1.353 \\
          &       & CosyPose~\cite{labbe2020cosypose} & -     & 0.480 & 0.606 & 0.812 & 0.633 & 0.550 \\
          &       & {Ours} & {-} & {0.432} & {0.560} & {0.761} & {0.584} & {0.274} \\
          &       & PVNet~\cite{SPeng18_PVNet} & -     & 0.428 & 0.543 & 0.754 & 0.575 & - \\
          &       & CDPN~\cite{li2019cdpn}  & -     & 0.393 & 0.537 & 0.779 & 0.569 & 0.279 \\
          &       & EPOS~\cite{hodan2020epos}  & -     & 0.389 & 0.501 & 0.750 & 0.547 & 0.468 \\
          &       & Pix2Pose~\cite{park2019pix2pose} & -     & 0.233 & 0.307 & 0.550 & 0.363 & 1.310 \\
    \midrule
    \multirow{2}[2]{*}{synt} & \multirow{2}[2]{*}{RGB} & EPOS~\cite{hodan2020epos}  & -     & 0.29  & 0.38  & 0.659 & 0.443 & 0.487 \\
          &       & DPOD~\cite{zakharov2019dpod}  & -     & 0.101 & 0.126 & 0.278 & 0.169 & 0.172 \\
    \midrule
    \multirow{3}[4]{*}{mix} & RGBD  & AAE~\cite{sundermeyer2018implicit}   & ICP   & 0.208 & 0.218 & 0.285 & 0.237 & 1.197 \\
\cmidrule{2-9}          & \multirow{2}[2]{*}{RGB} & Multi-Path AAE & -     & 0.15  & 0.153 & 0.346 & 0.217 & 0.200 \\
          &       & AAE~\cite{sundermeyer2018implicit}   & -     & 0.09  & 0.095 & 0.254 & 0.146 & 0.201 \\
    \midrule
    \multirow{2}[2]{*}{-} & \multirow{2}[2]{*}{depth} & Drost, PPF~\cite{drost2010model} & ICP   & 0.437 & 0.563 & 0.581 & 0.527 & 15.947$^*$ \\
          &       & Drost, PPF~\cite{drost2010model} & ICP, 3D edges & 0.409 & 0.525 & 0.542 & 0.492 & 3.389$^*$ \\
    \midrule
    \multirow{7}[3]{*}{real} & \multirow{4}[2]{*}{RGBD} & {Ours + Kabsch} & {ICP} & {0.557} & {0.739} & {0.773} & {0.69} & {0.393} \\
          &       & Ours + Kabsch & - & 0.557 & {0.721} & {0.759} & {0.679} & {0.329} \\
          &       & {Ours} & {ICP} & {0.478} & {0.62} & {0.66} & {0.586} & {0.453} \\
          &       & {Ours} & {-} & {0.403} & {0.553} & {0.595} & {0.517} & {0.343} \\
\cmidrule{2-9}          & \multirow{3}[1]{*}{RGB} & {Ours} & {4 calibrated views} & {0.598} & {0.751} & {0.793} & {0.714} & {1.326} \\
          &       & {Ours} & {2 calibrated  views} & {0.528} & {0.692} & {0.764} & {0.661} & {2.201} \\
          &       & {Ours} & {-} & {0.427} & {0.55} & {0.728} & {0.568} & {0.278} \\
    \end{tabular}%

    }
    
    \vspace{-2em}
    
\end{table*}%

\subsection{Results}

\textbf{Linemod.} We start this section by discussing the quantitative results of single object pose estimation on the Linemod dataset. Table~\ref{tab:lm_rgb_synt} and Table~\ref{tab:lm_rgb_real}  present the accuracy of pose estimation on monocular RGB images of methods trained on syhtetic and real data respectively. We report run times of each algorithm if the corresponding papers clearly state them. Run times are taken from the original papers, meaning that they are obtained using non-identical hardware. The table reports the percentages of correctly estimated poses w.r.t. the ADD score. The tables compare separately methods trained on synthetic data and methods trained on real data and methods with or without refinement. On synthetic data, the original DPOD reached the state of the art results beating the second best method almost by a factor of 2. The proposed approach further improved the results, surpassing DPOD by over 30\%, making the ADD score comparable even to ADD scores of some of the methods trained on real data. If real training data is used, the proposed approach reaches the best results among other methods trained on real data.  The proposed multi-view refiner clearly outperforms all other RGB-based refiners on the dataset even if only 2 views are used for refinement. It also proves that the refinement method handles symmetric object, such as Glue and Eggbox, well.

Table~\ref{tab:lm_rgbd_synt} and Table~\ref{tab:lm_rgbd_real} present quantitative results on Linemod datasets in case depth data is available for methods trained on synthetic and real data respectively. Only PPF, PointFusion, DenseFusion and our method use depth as input, all other methods use it only for pose refinement. Our detector easily performs on par or outperforms other methods even if no refinement is applied. It also beats the PPF-based solutions. Another important conclusion from these tables is relative independence of pose accuracy on whether synthetic or real data or which data modality was used for training. Due to the simplicity of the dataset and to the lack of occlusions, correspondences predicted from depth maps prove to be good enough for excellent pose accuracy if the Kabsch algorithm is used. In comparison do deep learning methods which utilize real train data~\cite{wang2019densefusion,xu2018pointfusion}, we did not explicitly rely on sensor fusion and used purely RGB or purely depth for correspondence estimation. This experiment also demonstrates the effectiveness of our proposed refiner, which easily reaches the performance scores of the best RGBD methods.

\textbf{Occlusion.} Moving to a more difficult scenario, where a varying number of objects is present in each image, we start evaluation with experiments on the Occlusion dataset. In this case, the ability to predict accurate segmentation masks and precise correspondences becomes of critical importance. The results for all types of detectors and training data types are summarized in Table~\ref{tab:lmo}. The table clearly shows that our method outperforms all other dense correspondence methods, e.g. CDPN, EPOS and Pix2Pose, while being only worse than CosyPose if PBR training data and no refinement were used. With the multi-view refinement our methods easily outperform all other RGBD methods and shows similar or slightly worse results than the top performing RGBD methods with ICP refinement. It also always improves the poses and the final score in contrast to ICP which worsened the AR score for some of the experiments.

Adding depth information and predicting segmentations and correspondences from depth rather than from RGB actually hurts the performance, as can be seen in the \textit{RGBD} block of the table. It indicates inferior quality of the predicted segmentations and correspondences, that are not sufficient for reliable and precise pose estimation. Additionaly, due to imprecise segmentations, Kabsch and ICP can potentially align the object to nearby objects. On the other hand, if depth is only used for the Kabsch algorithm with the correspondences predicted by the RGB correspondence network, a larger performance boost is observed. In this configuration, the proposed method sets the state-of-the-art results on the HBD dataset even without the ICP refinement. ICP boosts the results even further. This experiment and later the experiments on TLESS show that correspondence and segmentaton prediction from pure depth maps works well only on very simple datasets such as Linemod, where the object is not occluded and clearly stands out from the scene. With the more challenging datasets, RGB-based correspondence estimation starts to dominate due to richer RGB information available for the network.

\textbf{Homebrewed.} This dataset is the next step towards more complicated pose estimation benchmarks, as it consists of a larger number of more diverse objects and a varying degree of occlusions. The comparison is provided in Table~\ref{tab:hbd_bop}. Pose accuracy is reported in terms of Average Recall as defined in ~\cite{bopchallenge}. As indicated before in Table~\ref{tab:data_modalities}, we used exclusively synthetic PBR train data data. Our method easily outperforms all other RGB methods. With the multi-view refinement, it performs on par with the best RGBD methods. 

\setlength{\tabcolsep}{8pt}
\begin{table*}[!t]
\centering
  \centering
  \caption{\textbf{Pose estimation performance comparison on the Homebrewed dataset}: Results are reported in terms of the Average Recall score~\cite{bopchallenge}. The results prove the effectiveness of the proposed approach on all used data modalities. Run times are provided as they are reported in the BOP challenge~\cite{bopchallenge}  using non-identical hardware. PPF-based methods, labeled with $^*$ in the Time column, use only CPU. \label{tab:hbd_bop}}
  
  \resizebox{0.9\textwidth}{!}{%
  
        \begin{tabular}{c|c|c|c|ccc|c|c}
    \textbf{Train data} & \textbf{Data modality} & \textbf{Method} & \textbf{Refinement} & \textbf{VSD} & \textbf{MSSD} & \textbf{MSPD} & \textbf{AR} & \textbf{Time (s)} \\
    \midrule
    \multirow{15}[4]{*}{PBR} & \multirow{7}[2]{*}{RGBD} & {Ours + Kabsch} & {ICP} & {0.784} & {0.872} & {0.879} & {0.845} & {0.329} \\
          &       & {Ours + Kabsch} & {-} & {0.796} & {0.856} & {0.867} & {0.84} & {0.238} \\
          &       & CosyPose~\cite{labbe2020cosypose} & ICP   & 0.679 & 0.719 & 0.737 & 0.712 & 5.326 \\
          &       & CDPNv2~\cite{li2019cdpn} & ICP   & 0.629 & 0.757 & 0.749 & 0.712 & 0.713 \\
          &       & Pix2Pose~\cite{park2019pix2pose} & ICP   & 0.64  & 0.721 & 0.724 & 0.695 & 3.248 \\
          &       & {Ours} & {ICP} & {0.462} & {0.468} & {0.473} & {0.468} & {0.47} \\
          &       & {Ours} & {-} & {0.344} & {0.398} & {0.403} & {0.382} & {0.245} \\
\cmidrule{2-9}          & \multirow{8}[2]{*}{RGB} & {Ours} & {4 calibrated views} & {0.79} & {0.857} & {0.86} & {0.835} & {0.584} \\
          &       & {Ours} & {2 calibrated views} & {0.754} & {0.84} & {0.858} & {0.818} & {0.926} \\
          &       & CosyPose~\cite{labbe2020cosypose} & 8 uncalibrated views & 0.691 & 0.744 & 0.804 & 0.746 & 0.427 \\
          &       & {Ours} & {-} & {0.642} & {0.704} & {0.828} & {0.725} & {0.163} \\
          &       & CDPNv2~\cite{li2019cdpn} & -     & 0.614 & 0.708 & 0.845 & 0.722 & 0.273 \\
          &       & CosyPose~\cite{labbe2020cosypose} & 4 uncalibrated views & 0.646 & 0.685 & 0.756 & 0.696 & 0.445 \\
          &       & CosyPose~\cite{labbe2020cosypose} & -     & 0.613 & 0.634 & 0.721 & 0.656 & 	0.417 \\
          &       & EPOS~\cite{hodan2020epos}  & -     & 	0.484 & 0.527 & 0.729 & 0.58  & 0.657 \\
    \midrule
    \multirow{3}[4]{*}{synt} & RGBD  & Sundermeyer-IJCV19~\cite{sundermeyer2018implicit} & ICP   & 0.479 & 0.506 & 0.533 & 0.506 & 1.352 \\
\cmidrule{2-9}          & \multirow{2}[2]{*}{RGB} & Sundermeyer-IJCV19~\cite{sundermeyer2020augmented} & -     & 0.273 & 0.306 & 0.461 & 0.346 & 0.19 \\
          &       & DPOD~\cite{zakharov2019dpod}  & -     & 0.218 & 0.262 & 0.379 & 0.286 & 0.18 \\
    \midrule
    \multirow{4}[4]{*}{-} & \multirow{3}[2]{*}{Depth} & 	Vidal-Sensors18~\cite{vidal2018method} & ICP   & 0.707 & 0.704 & 0.708 & 0.706 & 2.608$^*$ \\
          &       & Drost-CVPR10-3D-Edges~\cite{drost2010model} & ICP, 3D edges & 0.618 & 0.624 & 0.626 & 0.623 & 127.372$^*$ \\
          &       & Drost-CVPR10-3D-Only~\cite{drost2010model} & ICP   & 0.593 & 0.627 & 0.627 & 0.615 & 16.136$^*$ \\
\cmidrule{2-9}          & RGBD  & Drost-CVPR10-Edges~\cite{drost2010model} & ICP,3D edges, images & 0.677 & 0.665 & 0.67  & 0.671 & 144.029$^*$ \\
    \bottomrule
    \end{tabular}%

    }
    \vspace{-1.5em}
\end{table*}

\begin{figure*}[!b]
\vspace{-2em}
\begin{subfigure}{.5\textwidth}
  \centering
  \includegraphics[width=.8\linewidth]{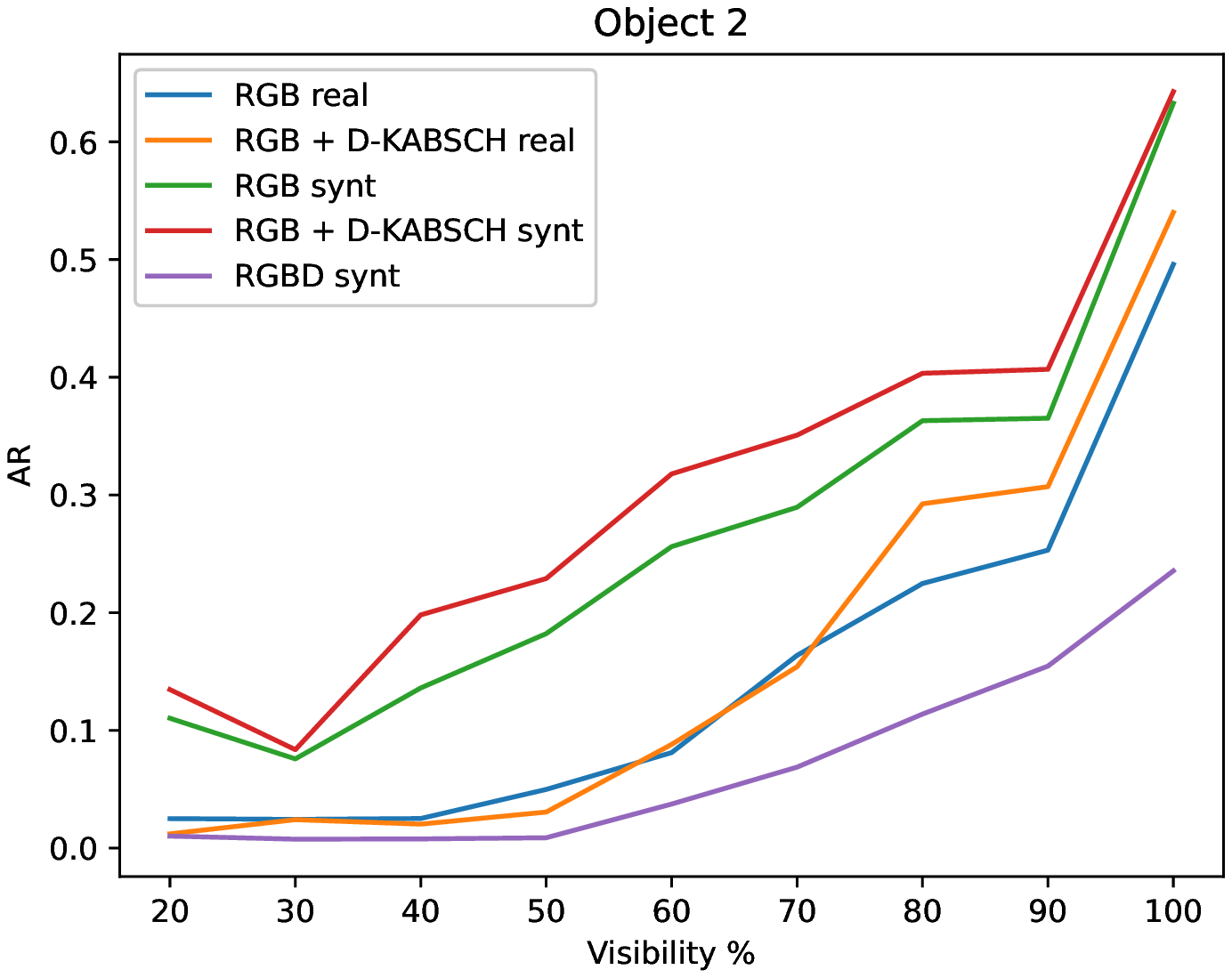}
\end{subfigure}%
\begin{subfigure}{.5\textwidth}
  \centering
  \includegraphics[width=.8\linewidth]{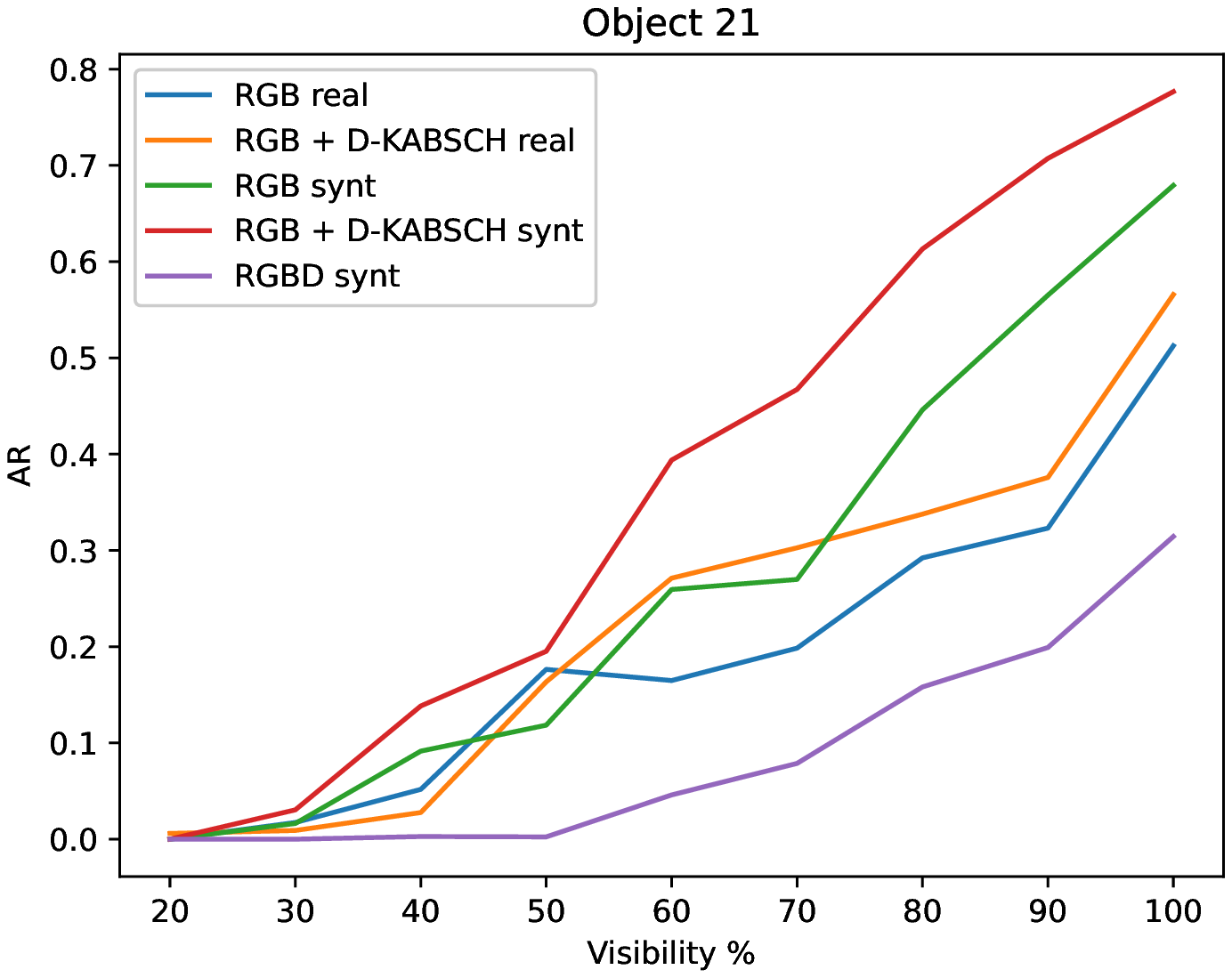}
\end{subfigure}

\caption{\textbf{Dependence of quality of pose estimation on object's visibility}. Pose quality is reported in terms of the Average Recall~\cite{bopchallenge} on objects 2 and 21 from the TLESS dataset. The plots show that correspondence prediction works from RGB works considerably more reliably than from depth in case of large occlusions. It also shows that synthetic training data allows for more reliable pose estimation of occluded objects. \label{fig:visinblity}}
\end{figure*}

\textbf{TLESS.} Table~\ref{tab:tless_bop} compares Average Recall of various configurations of the proposed detector to other participants of the BOP challenge 2020. Since TLESS dataset comes with a separate train set, we were able to test various combinations of YOLO and CENet. Our approach consistently outperforms all competing methods apart from CosyPose. It is interesting that usage of depth does improve the average recall as much as on other datasets. The lack of performance boost can be partially explained by the fact that approximately 16\% of images are taken by a camera looking almost parallel to the table with the objects. In this case, Primesense produces mostly empty depth maps for flat surfaces and empty on edges of objects, making pose prediction from occluded objects close to impossible. This also explains why in some cases AR score is lower if ICP refinement is applied. Again, the proposed multi-view refinement consistently improves the poses even in the presence of large occlusions and imperfect initial poses.

Another important insight from Table~\ref{tab:tless_bop} and from Table~\ref{tab:arr_ablation_tless} is that the correspondence estimation network trained on synthetic data actually outperforms the one trained on real data provided they are tested on exactly the same patches. There are several explanations to that. The real train data is very limited in the number of images and requires a lot of data augmentation to work on the real test set, while the synthetic data has considerably more images. Additionally, simulated synthetic train data naturally exhibits realistically looking occlusions, which helps generalize to the test images. It is also confirmed by the plot in Figure~\ref{fig:visinblity},  where CENet trained on synthetic data consistently outperforms the one trained on real data. If ground truth patches are used, the synthetic network achieves AR of 0.729, whereas the real network only 0.535. On the other hand, YOLO trained on real data outperforms YOLO train on synthetic data, which can be seen in much lower AR drop if the full real pipeline is used. As a result, we achieve the best result on TLESS if we combine a real YOLO and a synthetic correspondence estimation network, analogously to how it is done in AAE~\cite{sundermeyer2018implicit}.

\setlength{\tabcolsep}{8pt}
%TODO fix this table
\begin{table*}[!t]
\centering
  \centering
  \caption{\textbf{Pose estimation performance comparison on the BOP images of the TLESS dataset}: Results are reported in terms of the Average Recall score~\cite{bopchallenge}. The results prove the effectiveness of the proposed approach on all used data modalities. Run times are provided as they are reported in the BOP challenge~\cite{bopchallenge}  using non-identical hardware. PPF-based methods, labeled with $^*$ in the Time column, use only CPU. \label{tab:tless_bop}}
  
  \resizebox{0.9\textwidth}{!}{%
  
                  \begin{tabular}{c|c|c|c|ccc|c|c}
    \textbf{Train data} & \textbf{Data modality} & \textbf{Method} & \textbf{Refinement} & \textbf{VSD} & \textbf{MSSD} & \textbf{MSPD} & \textbf{AR} & \textbf{Time (s)} \\
    \midrule
    \multirow{11}[4]{*}{PBR} & \multirow{5}[2]{*}{RGBD} & {Ours + Kabsch} & {-} & {0.646} & {0.716} & {0.736} & \textbf{0.699} & {0.320} \\
          &       & {Ours + Kabsch} & {ICP} & {0.485} & {0.632} & {0.652} & {0.590} & {0.523} \\
          &       & CDPNv2~\cite{li2019cdpn} & ICP   & 0.368 & 0.449 & 0.488 & 0.435 & 2.486 \\
          &       & {Ours} & {ICP} & {0.277} & {0.320} & {0.339} & {0.312} & {0.688} \\
          &       & {Ours} & {-} & {0.251} & {0.303} & {0.312} & {0.289} & {0.330} \\
\cmidrule{2-9}          & \multirow{6}[2]{*}{RGB} & {Ours} & {2 calibrated views} & {0.634} & {0.707} & {0.728} & {0.690} & {1.293} \\
          &       & {Ours} & {4 calibrated views} & {0.645} & {0.710} & {0.712} & {0.689} & {0.886} \\
          &       & CosyPose~\cite{labbe2020cosypose} & -     & 0.571 & 0.589 & 0.761 & 0.64  & 0.493 \\
          &       & {Ours} & {-} & {0.561} & {0.602} & {0.744} & {0.636} & {0.328} \\
          &       & EPOS~\cite{hodan2020epos}  & -     & 0.38  & 0.403 & 0.619 & 0.467 & 1.992 \\
          &       & CDPNv2~\cite{li2019cdpn} & -     & 0.303 & 0.338 & 0.579 & 0.407 & 1.849 \\
    \midrule
    \multirow{2}[2]{*}{synt} & \multirow{2}[2]{*}{RGB} & EPOS~\cite{hodan2020epos}  & -     & 0.369 & 0.423 & 0.635 & 0.476 & 1.177 \\
          &       & DPOD~\cite{zakharov2019dpod}  & -     & 0.048 & 0.055 & 0.139 & 0.081 & 0.206 \\
    \midrule
    \multirow{4}[3]{*}{-} & \multirow{3}[2]{*}{Depth} & Vidal-Sensors18~\cite{vidal2018method} & ICP   & 0.464 & 0.575 & 0.574 & 0.538 & 7.063$^*$ \\
          &       & 	Drost-CVPR10-3D-Only~\cite{drost2010model} & ICP   & 0.375 & 0.478 & 0.480 & 0.444 & 9.204$^*$ \\
          &       & Drost-CVPR10-3D-Edges~\cite{drost2010model} & ICP, 3D edges & 0.370 & 0.422 & 0.420 & 0.404 & 62.507$^*$ \\
\cmidrule{2-9}          & \multirow{4}[1]{*}{RGBD}  & Drost-CVPR10-Edges~\cite{drost2010model} & ICP, 3D edges, images & 0.469 & 0.512 & 0.518 & 0.500 & 	70.914$^*$ \\
    \multirow{8}[3]{*}{real} &  & {Ours + Kabsch} & {-} & {0.537} & {0.557} & {0.586} & \textbf{0.56} & {0.315} \\
          &       & Pix2Pose~\cite{park2019pix2pose} & ICP   & 0.438 & 	0.548 & 0.549 & 0.512 & 4.180 \\
          &       & {Ours + Kabsch} & {ICP} & {0.416} & {0.516} & {0.537} & {0.490} & {0.578} \\
\cmidrule{2-9}          & \multirow{5}[2]{*}{RGB} & {Ours} & {4 calibrated views} & {0.467} & {0.541} & {0.563} & {0.524} & {0.727} \\
          &       & {Ours} & {2 calibrated views} & {0.461} & {0.530} & {0.578} & {0.523} & {0.973} \\
          &       & {Ours} & {-} & {0.469} & {0.491} & {0.595} & {0.518} & {0.367} \\
          &       & Pix2Pose~\cite{park2019pix2pose} & -     & 0.261 & 0.296 & 0.476 & 0.344 & 1.084 \\
          &       & Pix2Pose-Original-ICCV19~\cite{park2019pix2pose} & -     & 0.214 & 0.238 & 0.432 & 0.295 & 1.522 \\
    \midrule
    \multirow{14}[4]{*}{mix} & \multirow{9}[2]{*}{RGB} & CosyPose~\cite{labbe2020cosypose} & 8 uncalibrated views & 0.773 & 0.836 & 0.907 & 0.839 & 0.969 \\
          &       & CosyPose~\cite{labbe2020cosypose} & 4 uncalibrated views & 0.742 & 0.795 & 0.864 & 0.801 & 0.792 \\
          &       & CosyPose~\cite{labbe2020cosypose} & -     & 0.669 & 0.695 & 0.821 & 0.728 & 0.451 \\
          &       & {Ours} & {4 calibrated views} & {0.679} & {0.742} & {0.740} & {0.72} & {0.909} \\
          &       & {Ours} & {2 calibrated views} & {0.665} & {0.739} & {0.753} & {0.719} & {1.306} \\
          &       & {Ours} & {-} & {0.579} & {0.621} & {0.764} & {0.655} & {0.306} \\
          &       & CDPN~\cite{li2019cdpn}  & -     & 0.377 & 0.418 & 0.674 & 0.49  & 0.708 \\
          &       & CDPNv2~\cite{li2019cdpn} & -     & 0.386 & 0.426 & 0.62  & 0.478 & 	1.852 \\
          &       & Sundermeyer-IJCV19~\cite{sundermeyer2020augmented} & -     & 0.196 & 0.211 & 	0.504 & 0.304 & 0.194 \\
\cmidrule{2-9}          & \multirow{5}[2]{*}{RGBD} & {Ours + Kabsch} & {-} & {0.665} & {0.738} & {0.756} & {0.720} & 0.7367 \\
          &       & CosyPose~\cite{labbe2020cosypose} & ICP   & 0.587 & 0.749 & 0.767 & 0.701 & {0.274} \\
          &       & {Ours + Kabsch} & {ICP} & {0.503} & {0.656} & {0.672} & {0.610} & {0.455} \\
          &       & Sundermeyer-IJCV19~\cite{sundermeyer2020augmented} & ICP   & 0.459 & 0.489 & 0.514 & 0.487 & 0.531 \\
          &       & CDPNv2~\cite{li2019cdpn} & ICP   & 0.385 & 0.489 & 0.516 & 0.464 & 2.645 \\
    \bottomrule
    \end{tabular}%

    }
    
    \vspace{-2em}    
    
\end{table*}%

\subsection{Ablation Studies}
In the ablation studies, we measure what influences pose estimation the most, how imprecise YOLO detections affect the quality of correspondences and which data modality is better. Tables~\ref{tab:arr_ablation_lm}, \ref{tab:arr_ablation_lmo}, \ref{tab:arr_ablation_hbd} and \ref{tab:arr_ablation_tless}  report how pose estimation performance drops if YOLO detections are used instead of ground truth bounding boxes on Linemod and Occlusion, Homebrewed and TLESS datasets respectively. Those tables essentially report the upper bound of what the proposed approach can achieve, assuming perfect 2D detections, and how far from them the results actually are. They also additionally report quality of correspondences. In each object in the image, a correspondence quality is computed as median per-correspondence L2 error between a predicted correspondence and its correct ground truth location. Then, it is averaged across all objects. The metric takes into account object symmetries by choosing an equivalent symmetric pose.

\textbf{RGB vs Depth vs RGB + D-Kabsch.} With the advent of deep learning, much of pose estimation research has moved to monocular RGB~\cite{kehlSSD6DMakingRGBbased2017b,sundermeyer2018implicit,zakharov2019dpod,song2020hybridpose,labbe2020cosypose}. Even if depth information was used, it was used in a separate post-processing step for pose refinement and not for inference. In this paper, we introduced a pose estimation network capable of pose estimation from either modality, which leads to the question of which one is better. On the Linemod dataset (Table~\ref{tab:arr_ablation_lm}), RGB-only detectors still lag behind the depth-based counterparts especially if only synthetic data is used for their training.  It is also visible from the table that correspondences predicted from RGB tend to be better than those predicted from depth even for such a simplistic dataset. Table~\ref{tab:arr_ablation_lm} clearly shows the superior quality of correspondences estimated from RGB, Nevertheless, depth-based approaches still better than purely RGB approaches because of the simplicity of the dataset and the lack of occlusions.

On the other hand, the situation is different on more complicated datasets with occlusions and background clutter. On the Occlusion dataset, the RGB version of CENet works better than the depth version. Depth brings improvement only if the correspondences are computed from RGB images and then projected onto the point cloud for the Kabsch algorithm. This is also confirmed by the correspondence error analysis in Table~\ref{tab:arr_ablation_lmo}, where correspondences computed from RGB are almost two times more precise than the once computed from depth maps. This trend is also noticeable on the more complicated Occlusion dataset (Table~\ref{tab:arr_ablation_lmo}). It is clear from the table that the correspondence error is much larger due to occlusions.  As seen from Table~\ref{tab:hbd_bop}, on the Homebrewed dataset the proposed approach trained on RGB only significantly outperforms its counterpart trained on depth, while the RGB variant with the Kabsch algorithm shows outstanding AR. That is explained by less accurate predictions of segmentation masks and correspondences for occluded objects when only depth information is used for that. Table~\ref{tab:arr_ablation_hbd} clearly shows that correspondences predicted from RGB are almost five times better then their depth-based counterparts, which explains the difference in pose estimation performance. On TLESS dataset, as seen from Table~\ref{tab:tless_bop} and Table~\ref{tab:arr_ablation_tless}, using depth again does not help at all as the quality of correspondences deteriorate. Depth CENet also produced visually worse segmentations. This phenomenon can be explained by larger degrees of occlusion and camera angles resulting in missing or very noisy depth values. Figure~\ref{fig:visinblity}, additionally proves strong dependence of pose estimation quality from depth maps on the visibility of objects. It show that RGB-based CENet is much better capable of handling occlusions that the depth-based CENet.

To sum up, semantic segmentation and correspondence estimation is considerably more precise when done completely from RGB. If depth information is present, it is more beneficial to still predict them from RGB and then project into the 3D space and than use the Kabsch algorithm.  Such an approach potentially also allows for alternating between PnP and Kabsch based on the quality of depth maps.

\setlength{\tabcolsep}{5pt}
\begin{table}[!t]
  \centering
  \caption{\textbf{Pose estimation performance on Linemod on different data modalities and types of train data}: The table reports the percentages of correctly estimated poses w.r.t. the ADD score. An ADD score for each data modality and train data type is provided separately on ground truth 2D detections and on YOLO detections. Symmetry-aware median $L_2$ correspondence error demonstrates the quality of predicted correspondences.\label{tab:arr_ablation_lm}}
  
  \resizebox{1\linewidth}{!}{%
  
        \begin{tabular}{ccccc|c|c}
    Dataset & Data modality & Data type & Bboxes & \multicolumn{1}{c}{Refinement} & \multicolumn{1}{c}{ADD} & Corr. Err. \\
    \midrule
    \multirow{28}[24]{*}{Linemod} & \multirow{12}[8]{*}{RGB} & \multirow{6}[4]{*}{real} & \multirow{3}[2]{*}{GT} & -     & 93.99 & \multirow{3}[2]{*}{2.48} \\
          &       &       &       & 2 views & 97.42 &  \\
          &       &       &       & 4 views & 99.79 &  \\
\cmidrule{4-7}          &       &       & \multirow{3}[2]{*}{YOLO} & -     & 93.59 & \multirow{3}[2]{*}{2.52} \\
          &       &       &       & 2 views & 97.68 &  \\
          &       &       &       & 4 views & 99.91 &  \\
\cmidrule{3-7}          &       & \multirow{6}[4]{*}{synt} & \multirow{3}[2]{*}{GT} & -     & 81.20 & \multirow{3}[2]{*}{4.61} \\
          &       &       &       & 2 views & 96.47 &  \\
          &       &       &       & 4 views & 99.63 &  \\
\cmidrule{4-7}          &       &       & \multirow{3}[2]{*}{YOLO} & -     & 80.39 & \multirow{3}[2]{*}{4.67} \\
          &       &       &       & 2 views & 96.33 &  \\
          &       &       &       & 4 views & 99.70 &  \\
\cmidrule{2-7}          & \multirow{8}[8]{*}{RGBD} & \multirow{4}[4]{*}{real} & \multirow{2}[2]{*}{GT} & -     & 99.85 & \multirow{2}[2]{*}{5.31} \\
          &       &       &       & ICP   & 99.85 &  \\
\cmidrule{4-7}          &       &       & \multirow{2}[2]{*}{YOLO} & -     & 98.81 & \multirow{2}[2]{*}{5.38} \\
          &       &       &       & ICP   & 99.13 &  \\
\cmidrule{3-7}          &       & \multirow{4}[4]{*}{synt} & \multirow{2}[2]{*}{GT} & -     & 97.99 & \multirow{2}[2]{*}{15.81} \\
          &       &       &       & ICP   & 98.19 &  \\
\cmidrule{4-7}          &       &       & \multirow{2}[2]{*}{YOLO} & -     & 88.94 & \multirow{2}[2]{*}{17.32} \\
          &       &       &       & ICP   & 90.06 &  \\
\cmidrule{2-7}          & \multirow{8}[8]{*}{RGB + D-Kabsch} & \multirow{4}[4]{*}{real} & \multirow{2}[2]{*}{GT} & -     & 99.99 & \multirow{2}[2]{*}{2.48} \\
          &       &       &       & ICP   & 99.99 &  \\
\cmidrule{4-7}          &       &       & \multirow{2}[2]{*}{YOLO} & -     & 99.73 & \multirow{2}[2]{*}{2.52} \\
          &       &       &       & ICP   & 99.64 &  \\
\cmidrule{3-7}          &       & \multirow{4}[4]{*}{synt} & \multirow{2}[2]{*}{GT} & -     & 99.94 & \multirow{2}[2]{*}{4.61} \\
          &       &       &       & ICP   & 99.97 &  \\
\cmidrule{4-7}          &       &       & \multirow{2}[2]{*}{YOLO } & -     & 99.22 & \multirow{2}[2]{*}{4.67} \\
          &       &       &       & ICP   & 99.18 &  \\
    \bottomrule
    \end{tabular}%
    
    }
    
    \vspace{-2em}

\end{table}%

\textbf{Real vs Synthetic Data.} As an experiment from~\cite{kaskman2019homebreweddb} showed, training pose estimation networks on real RGB data tends to outperformed the one on synthetic if images and poses come from exactly the same domain. That is still clearly visible on Linemod dataset in Table~\ref{tab:lm_rgb_synt}. Detectors trained on synthetic data are still behind their real counterparts even in spite of the progress in synthetic data generation. It is also visible from the worse quality of correspondences as seen in Table~\ref{tab:arr_ablation_lm}. The same holds for CENet train on synthetic depth data.  On the other hand, as can be seen from TLESS Tables~\ref{tab:tless_bop} and~\ref{tab:arr_ablation_tless} YOLO benefits a lot from training on real data even if the train data is out of test set domain. At the same time, CENet trained on synthetic data clearly outperforms the CENet trained on real data both in terms of AR and the quality of correspondences due to better pose space coverage. 

To sum up, real train data is better for training 2D detectors. For training correspondence estimation network, synthetic data is beneficial unless train and test data come from the same domain and contain objects in very similar poses.

 \begin{figure*}[!t]
\centering

\begin{subfigure}{.38\textwidth}
  \centering
  
  \hfill
  \includegraphics[width=.45\linewidth]{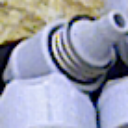}
  \hfill
  \includegraphics[width=.45\linewidth]{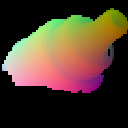}
  \hfill
  
  \hphantom{}
   
  \hfill
  \subfloat[Input patch]{
  \includegraphics[width=.45\linewidth]{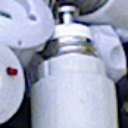}
  }
  \hfill
  \subfloat[GT NOCS]{
  \includegraphics[width=.45\linewidth]{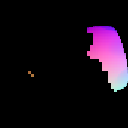}
  }
  \hfill
  
\end{subfigure}%
\begin{subfigure}{.38\textwidth}
  \centering
  
     \hfill
   \includegraphics[width=.45\linewidth]{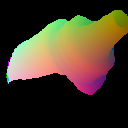}
  \hfill
  \includegraphics[width=.45\linewidth]{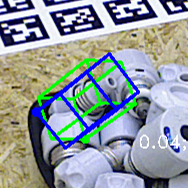}
  \hfill
  
  \hphantom{}
  
  \hfill
  \subfloat[Predicted NOCS]{
   \includegraphics[width=.45\linewidth]{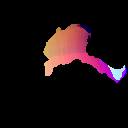}
   }
  \hfill
  \subfloat[Predicted pose]{
  \includegraphics[width=.45\linewidth]{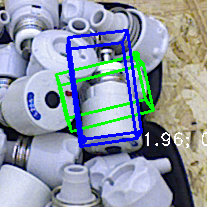}
  }
  \hfill
  
\end{subfigure}

\caption{\textbf{Success cases and failure cases.} Top row provides an example of successfully estimated segmentation mask and NOCS correspondences even in case of occlusions and similar objects present in the image patch. The bottom row illustrates a failure case, where the CENet is confused by occlusions and similar objects belonging to other object classes, which leads to an incorrect estimated pose.  The green cuboid represents the ground truth pose (up to a symmetry transformation), whereas the blue cuboid represents the estimated pose.\label{fig:success_failures}}
\vspace{-1.5em}
\end{figure*}

\setlength{\tabcolsep}{5pt}
\begin{table}[!t]
  \centering
  \caption{\textbf{Pose estimation performance on the Occlusion dataset  on different data modalities and type of train data}: The table reports in terms of the AR score~\cite{bopchallenge}. An AR score for each data modality and train data type is provided separately on ground truth 2D detections and on YOLO detections. Symmetry-aware median $L_2$ correspondence error demonstrates the quality of predicted correspondences.\label{tab:arr_ablation_lmo}}
  
  \resizebox{1\linewidth}{!}{%
  
        \begin{tabular}{c|cccccc}
    \textbf{Dataset} & \textbf{Data modality} & \textbf{Data type} & \textbf{Bboxes} & \textbf{Refinement} & \textbf{AR} & \textbf{Corr. Err.} \\
    \midrule
    \multirow{28}[24]{*}{LMO} & \multirow{12}[8]{*}{RGB} & \multirow{6}[4]{*}{real} & \multirow{3}[2]{*}{GT} & -     & 0.601 & \multirow{3}[2]{*}{13.830} \\
          &       &       &       & 2 views & 0.690 &  \\
          &       &       &       & 4 views & 0.751 &  \\
\cmidrule{4-7}          &       &       & \multirow{3}[2]{*}{YOLO } & -     & 0.568 & \multirow{3}[2]{*}{16.150} \\
          &       &       &       & 2 views & 0.661 &  \\
          &       &       &       & 4 views & 0.714 &  \\
\cmidrule{3-7}          &       & \multirow{6}[4]{*}{synt} & \multirow{3}[2]{*}{GT} & -     & 0.649 & \multirow{3}[2]{*}{12.275} \\
          &       &       &       & 2 views & 0.741 &  \\
          &       &       &       & 4 views & 0.773 &  \\
\cmidrule{4-7}          &       &       & \multirow{3}[2]{*}{YOLO } & -     & 0.584 & \multirow{3}[2]{*}{12.987} \\
          &       &       &       & 2 views & 0.660 &  \\
          &       &       &       & 4 views & 0.695 &  \\
\cmidrule{2-7}          & \multirow{8}[8]{*}{RGBD} & \multirow{4}[4]{*}{real} & \multirow{2}[2]{*}{GT} & -     & 0.534 & \multirow{2}[2]{*}{26.482} \\
          &       &       &       & ICP   & 0.609 &  \\
\cmidrule{4-7}          &       &       & \multirow{2}[2]{*}{YOLO} & -     & 0.517 & \multirow{2}[2]{*}{26.611} \\
          &       &       &       & ICP   & 0.586 &  \\
\cmidrule{3-7}          &       & \multirow{4}[4]{*}{synt} & \multirow{2}[2]{*}{GT} & -     & 0.606 & \multirow{2}[2]{*}{21.267} \\
          &       &       &       & ICP   & 0.657 &  \\
\cmidrule{4-7}          &       &       & \multirow{2}[2]{*}{YOLO} & -     & 0.582 & \multirow{2}[2]{*}{22.356} \\
          &       &       &       & ICP   & 0.541 &  \\
\cmidrule{2-7}          & \multirow{8}[8]{*}{RGB + D-Kabsch} & \multirow{4}[4]{*}{real} & \multirow{2}[2]{*}{GT} & -     & 0.738 & \multirow{2}[2]{*}{13.830} \\
          &       &       &       & ICP   & 0.757 &  \\
\cmidrule{4-7}          &       &       & \multirow{2}[2]{*}{YOLO} & -     & 0.679 & \multirow{2}[2]{*}{16.150} \\
          &       &       &       & ICP   & 0.690 &  \\
\cmidrule{3-7}          &       & \multirow{4}[4]{*}{synt} & \multirow{2}[2]{*}{GT} & -     & 0.796 & \multirow{2}[2]{*}{12.987} \\
          &       &       &       & ICP   & 0.793 &  \\
\cmidrule{4-7}          &       &       & \multirow{2}[2]{*}{YOLO } & -     & 0.582 & \multirow{2}[2]{*}{22.356} \\
          &       &       &       & ICP   & 0.698 &  \\
    \bottomrule
    \end{tabular}%
    
    }
    
    \vspace{-2em}

\end{table}%

\textbf{Choice of PnP and RANSAC}. In all the reported RGB results we relied on the standard implementation of EPnP~\cite{lepetit2009epnp} and of RANSAC from OpenCV, although uncountable number of improved versions of PnP and RANSAC have been published in the past~\cite{raguram2012usac,barath2018graph,chum2003locally}. Additionally, EPOS~\cite{hodan2020epos} reported, that a large improvement can be achieved by simple switching from OpenCV PnP+RANSAC to Graph Cut RANSAC~\cite{barath2018graph} with DLS-PnP~\cite{hesch2011direct}. We have tried this combination on scene 3 of TLESS. It resulted in increase of AR from 21.45 to, at most, 24.02. However, we faced two complications. First, due to a large number of correspondences, RANSAC takes over a second per each object, making the entire pipeline slow and not applicable in practice. Secondly, the algorithm seemed to be very sensitive to hyper-parameters, and optimal hyper-parameters are not shared among the objects. As a result, OpenCV implementation proved to be more rapid and robust, even tough potentially resulting in sub-optimal AR scores.

\textbf{Limitations of CENet}. There are two main limiting factors of the CENet. The first comes from the fact that the reliable pose estimation with correspondences requires their good quality and the right number of them to deal with outliers. Therefore, large occlusions might be problematic, as indicated in Figure~\ref{fig:visinblity} and Figure~\ref{fig:success_failures}. However, pose estimation of occluded object is a common problem for all methods. Another issue, illustrated in Figure~\ref{fig:visinblity} , arises when CENet receives a patch containing similarly looking objects or object of the same class.

\textbf{Multi-View Refinement} We have experimented with the multi-view refinement on all datasets, and on all datasets it proved to be effective, as opposed to ICP which did not always improve the poses. It can be explained by its independence from the quality of depth data and potentially better handling of objects which are occluded in only some of the frames used for refinement. We split each sequence into non-overlapping sets of 2 or 4 images used for refinement. The pose is optimized jointly and then transformed into the coordinate system of each frame. Even though the number of frames is lower than in CosyPose~\cite{labbe2020cosypose}, the proposed refiner anyway produces better or similar relative improvement. This is also explained by the fact that CosyPose optimized camera and object poses jointly while we assume the availability of relative camera poses. The CosyPose refiner performs noticeably better on TLESS, but this can be explained by a much better initial poses produced by CosyPose.  For each set, we compute relative camera poses from individual ground truth camera poses.  In these experiments, the images were split into batches absolutely at random. The analysis of how a choice of the camera samplig strategy affects the final score is provided in the supplementary.

%% file: conclusion.tex
\section{Conclusions}

In this paper, we extend the Dense Pose Object Detector (DPOD) method, by splitting its inference into three stages with a optional fourth refinement stage. In the first stage, YOLO outputs 2D bounding boxes of the objects of interest. In the second stage, our unified CENet network predicts foreground object masks and dense 2D-3D correspondences between image pixels and corresponding 3D models either from RGB or depth input modalities with the same architecture. Object's 6 DoF is then reliably estimated with PnP+RANSAC or with Kabsch+RANSAC depending on the available data modality in the third stage. In the optional fourth stage, predicted poses are refined either with ICP or with the proposed novel multi-view refiner. The second stage network can be trained both on RGB images and on depth masks. We additionally proposed a new multi-view pose refiner based on differentiable rendering, which is used to produce a pose which is globally consistent with dense NOCS correspondences predicted in all frames.  The proposed approaches have been tested on four popular datasets, each of which is challenging in its own way: Linemod, Occlusion, Homebrewed and TLESS. The detector shows excellent results on all the datasets and data modalities while still staying fast and scalable. The refiner improves the poses even further while still being reasonably fast. We provide an extensive evaluation and measured effects of the choice of data modality, and of the choice of the training data: synthetic or real. The overall results indicated that RGB is good for correspondence prediction, while depth is good for pose prediction. The more occlusions we have in the data the more benefits we get from the dense correspondences estimated from RGB. In that case, noisy depth still improves the pose with respect to the pose estimated from RGB only. 

\setlength{\tabcolsep}{5pt}
\begin{table}[h]
  \centering
  \caption{\textbf{Pose estimation performance on the Homebrewed  on different data modalities and type of train data}: The table reports the percentages of correctly estimated poses w.r.t. the Average Recall score~\cite{bopchallenge}. An AR score for each data modality and train data type is provided separately on ground truth 2D detections and on YOLO detections. Symmetry-aware median $L_2$ correspondence error demonstrates the quality of predicted correspondences.\label{tab:arr_ablation_hbd}}
  
  \resizebox{1\linewidth}{!}{%
  
        \begin{tabular}{c|cccc|c|c}
    \textbf{Dataset} & \textbf{Data modality} & \textbf{Data type} & \textbf{Bboxes} & \multicolumn{1}{c}{\textbf{Refinement}} & \multicolumn{1}{c}{\textbf{AR}} & \textbf{Corr. Err.} \\
    \midrule
    \multirow{14}[12]{*}{\textbf{HBD}} & \multirow{6}[4]{*}{RGB} & \multirow{6}[4]{*}{synt} & \multirow{3}[2]{*}{GT} & -     & 0.723 & \multirow{3}[2]{*}{9.52} \\
          &       &       &       & 2 views & 0.808 &  \\
          &       &       &       & 4 views & 0.807 &  \\
\cmidrule{4-7}          &       &       & \multirow{3}[2]{*}{YOLO } & -     & 0.668 & \multirow{3}[2]{*}{13.78} \\
          &       &       &       & 2 views & 0.766 &  \\
          &       &       &       & 4 views & 0.783 &  \\
\cmidrule{2-7}          & \multirow{4}[4]{*}{RGBD} & \multirow{4}[4]{*}{synt} & \multirow{2}[2]{*}{GT} & -     & 0.375 & \multirow{2}[2]{*}{52.69} \\
          &       &       &       & ICP   & 0.453 &  \\
\cmidrule{4-7}          &       &       & \multirow{2}[2]{*}{YOLO} & -     & 0.362 & \multirow{2}[2]{*}{54.10} \\
          &       &       &       & ICP   & 0.438 &  \\
\cmidrule{2-7}          & \multirow{4}[4]{*}{RGB + D-Kabsch} & \multirow{4}[4]{*}{synt} & \multirow{2}[2]{*}{GT} & -     & 0.843 & \multirow{2}[2]{*}{9.52} \\
          &       &       &       & ICP   & 0.853 &  \\
\cmidrule{4-7}          &       &       & \multirow{2}[2]{*}{YOLO } & -     & 0.793 & \multirow{2}[2]{*}{13.78} \\
          &       &       &       & ICP   & 0.801 &  \\
    \bottomrule
    \end{tabular}%
    
    }

\end{table}%

\setlength{\tabcolsep}{5pt}
\begin{table}[!t]
  \centering
  \caption{\textbf{Pose estimation performance on the TLESS  on different data modalities and type of train data}: The table reports the percentages of correctly estimated poses w.r.t. the Average Recall score~\cite{bopchallenge}. An AR score for each data modality and train data type is provided separately on ground truth 2D detections and on YOLO detections. Symmetry-aware median $L_2$ correspondence error demonstrates the quality of predicted correspondences.\label{tab:arr_ablation_tless}}
  
  \resizebox{1\linewidth}{!}{%
        \begin{tabular}{c|cccc|c|c}
    \textbf{Dataset} & \textbf{Data modality} & \textbf{Data type} & \textbf{Bboxes} & \textbf{Refinement} & \textbf{AR} & \textbf{Corr. Err.} \\
    \midrule
    \multirow{29}[24]{*}{\textbf{TLESS}} & \multirow{15}[10]{*}{RGB} & \multirow{6}[4]{*}{real} & \multirow{3}[2]{*}{GT} & -     & 0.535 & \multirow{3}[2]{*}{26.946} \\
          &       &       &       & 2 views & 0.542 &  \\
          &       &       &       & 4 views & 0.548 &  \\
\cmidrule{4-7}          &       &       & \multirow{3}[2]{*}{YOLO } & -     & 0.518 & \multirow{3}[2]{*}{27.516} \\
          &       &       &       & 2 views & 0.523 &  \\
          &       &       &       & 4 views & 0.524 &  \\
\cmidrule{3-7}          &       & \multirow{6}[4]{*}{synt} & \multirow{3}[2]{*}{GT} & -     & 0.729 & \multirow{3}[2]{*}{10.400} \\
          &       &       &       & 2 views & 0.798 &  \\
          &       &       &       & 4 views & 0.795 &  \\
\cmidrule{4-7}          &       &       & \multirow{3}[2]{*}{YOLO } & -     & 0.636 & \multirow{3}[2]{*}{20.616} \\
          &       &       &       & 2 views & 0.689 &  \\
          &       &       &       & 4 views & 0.690 &  \\
\cmidrule{3-7}          &       & \multirow{3}[2]{*}{mix} & \multirow{3}[2]{*}{YOLO } & -     & 0.655 & \multirow{3}[2]{*}{18.373} \\
          &       &       &       & 2 views & 0.719 &  \\
          &       &       &       & 4 views & 0.720 &  \\
\cmidrule{2-7}          & \multirow{4}[4]{*}{RGBD} & \multirow{4}[4]{*}{synt} & \multirow{2}[2]{*}{GT} & -     & 0.322 & \multirow{2}[2]{*}{34.396} \\
          &       &       &       & ICP   & 0.354 &  \\
\cmidrule{4-7}          &       &       & \multirow{2}[2]{*}{YOLO} & -     & 0.289 & \multirow{2}[2]{*}{41.087} \\
          &       &       &       & ICP   & 0.312 &  \\
\cmidrule{2-7}          & \multirow{10}[10]{*}{RGB + D-Kabsch} & \multirow{4}[4]{*}{real} & \multirow{2}[2]{*}{GT} & -     & 0.583 & \multirow{2}[2]{*}{26.946} \\
          &       &       &       & ICP   & 0.513 &  \\
\cmidrule{4-7}          &       &       & \multirow{2}[2]{*}{YOLO} & -     & 0.560 & \multirow{2}[2]{*}{27.516} \\
          &       &       &       & ICP   & 0.490 &  \\
\cmidrule{3-7}          &       & \multirow{4}[4]{*}{synt} & \multirow{2}[2]{*}{GT} & -     & 0.812 & \multirow{2}[2]{*}{10.400} \\
          &       &       &       & ICP   & 0.678 &  \\
\cmidrule{4-7}          &       &       & \multirow{2}[2]{*}{YOLO } & -     & 0.699 & \multirow{2}[2]{*}{20.616} \\
          &       &       &       & ICP   & 0.590 &  \\
\cmidrule{3-7}          &       & \multirow{2}[2]{*}{mix} & \multirow{2}[2]{*}{YOLO } & -     & 0.720 & \multirow{2}[2]{*}{18.373} \\
          &       &       &       & ICP   & 0.610 &  \\
    \bottomrule
    \end{tabular}%
    
    }
    \vspace{-1em}

\end{table}%

%% file: supp_arxiv.tex
\section{Supplementary Material}
\textbf{Frame selection strategy}  In this experiment, we tested how selection of the frames for joint refinement affects the overall performence of the refiner. Tables~\ref{tab:refiner_lm},\ref{tab:refiner_lmo},\ref{tab:refiner_hbd},\ref{tab:refiner_tless} summarize these experiments. We experimented with three different strategies: closest view sampling, random sampling and furthest view sampling. Closest view sampling corresponds to the worst case scenario because it does not provide strong enough geometric constraints. In other words, an error in pose estimation from the reference frame might not be necessarily be visible in the other frames used for refinement. Occlusions will cause further problems.

\begin{table}[!b]
  \centering
  \caption{Different view sampling strategies for the multi-view refiner on the Linemod~\cite{hinterstoisser2012model} dataset.\label{tab:refiner_lm}}
    \begin{tabular}{cccc|c}
    Dataset & Data type & N views & Views sampling & ADD \\
    \midrule
    \multirow{14}[12]{*}{Linemod} & \multirow{7}[6]{*}{real} & -     & -     & 93.59 \\
\cmidrule{3-5}          &       & \multirow{3}[2]{*}{2 views} & Closest & 86.43 \\
          &       &       & Random & 99.52 \\
          &       &       & Furthest  & 99.71 \\
\cmidrule{3-5}          &       & \multirow{3}[2]{*}{4 views} & Closest & 96.53 \\
          &       &       & Random & 99.91 \\
          &       &       & Furthest  & 99.97 \\
\cmidrule{2-5}          & \multirow{7}[6]{*}{synt} & -     & -     & 81.20 \\
\cmidrule{3-5}          &       & \multirow{3}[2]{*}{2 views} & Closest & 87.45 \\
          &       &       & Random & 98.95 \\
          &       &       & Furthest  & 99.02 \\
\cmidrule{3-5}          &       & \multirow{3}[2]{*}{4 views} & Closest & 94.65 \\
          &       &       & Random & 99.70 \\
          &       &       & Furthest  & 99.80 \\
    \bottomrule
    \end{tabular}%
  
\end{table}%

As expected, the closest view sampling performs the worst among these strategies, but poses still improve in most of the cases. Random and furthest sampling tend to perform similarly to each other. This is explained that furthest views do not necessarily correspond to the best possible configuration, as the error might not be visible in them. Additionally, furthest view sampling might result in selecting views which are too far from the reference frame, introducing new occlusions.

\begin{table}[!t]
  \centering
  \caption{Different view sampling strategies for the multi-view refiner on the Occlusion~\cite{brachmannUncertaintyDriven6DPose2016a} dataset. \label{tab:refiner_lmo}}
    \begin{tabular}{cccc|c}
    Dataset & Data type & N views & Views sampling & AR \\
    \midrule
    \multirow{14}[12]{*}{LMO} & \multirow{7}[6]{*}{real} & -     & -     & 0.568 \\
\cmidrule{3-5}          &       & \multirow{3}[2]{*}{2 views} & Closest & 0.627 \\
          &       &       & Random & 0.661 \\
          &       &       & Furthest  & 0.648 \\
\cmidrule{3-5}          &       & \multirow{3}[2]{*}{4 views} & Closest & 0.693 \\
          &       &       & Random & 0.714 \\
          &       &       & Furthest  & 0.706 \\
\cmidrule{2-5}          & \multirow{7}[6]{*}{synt} & -     & -     & 0.584 \\
\cmidrule{3-5}          &       & \multirow{3}[2]{*}{2 views} & Closest & 0.627 \\
          &       &       & Random & 0.660 \\
          &       &       & Furthest  & 0.664 \\
\cmidrule{3-5}          &       & \multirow{3}[2]{*}{4 views} & Closest & 0.693 \\
          &       &       & Random & 0.695 \\
          &       &       & Furthest  & 0.687 \\
    \bottomrule
    \end{tabular}%
\end{table}%

\begin{table}[t]
  \centering
  \caption{Different view sampling strategies for the multi-view refiner on the Homebrewed~\cite{kaskman2019homebreweddb} dataset. \label{tab:refiner_hbd}}
    \begin{tabular}{cccc|c}
    Dataset & Data type & N views & Views sampling & AR \\
    \midrule
    \multirow{7}[6]{*}{HBD} & \multirow{7}[6]{*}{synt} & -     & -     & 0.668 \\
\cmidrule{3-5}          &       & \multirow{3}[2]{*}{2 views} & Closest & 0.693 \\
          &       &       & Random & 0.808 \\
          &       &       & Furthest  & 0.762 \\
\cmidrule{3-5}          &       & \multirow{3}[2]{*}{4 views} & Closest & 0.724 \\
          &       &       & Random & 0.783 \\
          &       &       & Furthest  & 0.775 \\
    \bottomrule
    \end{tabular}%
\end{table}%

\begin{table}[t]
  \centering
  \caption{Different view sampling strategies for the multi-view refiner on the TLESS~\cite{hodan2017tless} dataset. \label{tab:refiner_tless}}
    \begin{tabular}{cccc|c}
    Dataset & Data type & N views & Views sampling & AR \\
    \midrule
    \multirow{21}[18]{*}{TLESS} & \multirow{7}[6]{*}{real} & -     & -     & 0.518 \\
\cmidrule{3-5}          &       & \multirow{3}[2]{*}{2 views} & Closest & 0.524 \\
          &       &       & Random & 0.523 \\
          &       &       & Furthest  & 0.534 \\
\cmidrule{3-5}          &       & \multirow{3}[2]{*}{4 views} & Closest & 0.544 \\
          &       &       & Random & 0.524 \\
          &       &       & Furthest  & 0.540 \\
\cmidrule{2-5}          & \multirow{7}[6]{*}{synt} & -     & -     & 0.636 \\
\cmidrule{3-5}          &       & \multirow{3}[2]{*}{2 views} & Closest & 0.693 \\
          &       &       & Random & 0.690 \\
          &       &       & Furthest  & 0.694 \\
\cmidrule{3-5}          &       & \multirow{3}[2]{*}{4 views} & Closest & 0.712 \\
          &       &       & Random & 0.689 \\
          &       &       & Furthest  & 0.707 \\
\cmidrule{2-5}          & \multirow{7}[6]{*}{mix} & -     & -     & 0.655 \\
\cmidrule{3-5}          &       & \multirow{3}[2]{*}{2 views} & Closest & 0.715 \\
          &       &       & Random & 0.719 \\
          &       &       & Furthest  & 0.725 \\
\cmidrule{3-5}          &       & \multirow{3}[2]{*}{4 views} & Closest & 0.738 \\
          &       &       & Random & 0.720 \\
          &       &       & Furthest  & 0.742 \\
    \bottomrule
    \end{tabular}%
\end{table}%